\documentclass{article}
\usepackage{stywhispers,amsmath,amsfonts,epsfig}
\usepackage{graphicx}
\usepackage{subcaption}
\usepackage{hyperref}

\def\x{{\mathbf x}}
\def\u{{\mathbf u}}
\def\v{{\mathbf v}}
\def\m{{\mathbf m}}
\def\R{{\mathbb R}}

\title{Exploring the high dimensional geometry of HSI features}
\name{Wojciech Czaja$^{1,2}$\thanks{Wojciech Czaja acknowledges support from the NSF under the DMS grant \# 1738003 and from the ARO under the grant \# W911NF-17-1-0304.}, Ilya Kavalerov$^3$\thanks{This work was supported in part by the MURI from the Army Research Office under the Grant No. W911NF-17-1-0304. This is part of the collaboration between US DOD, UK MOD and UK Engineering and Physical Research Council (EPSRC) under the Multidisciplinary University Research Initiative.}, Weilin Li$^4$\thanks{Welin Li thanks the AMS-Simons Travel Grant for support.}}
\address{$^1$Department of Mathematics, University of Maryland, College Park \\ $^2$ Center for Scientific Computation and Mathematical Modeling, University of Maryland, College Park \\ $^3$UMIACS, University of Maryland, College Park \\ $^4$Courant Institute of Mathematical Sciences, New York University}

\begin{document}
%
\maketitle
\begin{abstract}
We explore feature space geometries induced by the 3-D Fourier scattering transform and deep neural network with extended attribute profiles on four standard hyperspectral images. We examine the distances and angles of class means, the variability of classes, and their low-dimensional structures. These statistics are compared to that of raw features, and our results provide insight into the vastly different properties of these two methods. We also explore a connection with the newly observed deep learning phenomenon of neural collapse. 
\end{abstract}

\section{Introduction}
\label{sec:intro}
\vskip-0.2cm
Some of the fundamental methods in pattern recognition and machine learning are based on simple geometric notions. Support vector machines split data points into two classes by a maximum margin hyperplane, $k$-means clustering attempts to group points into isotropic balls, and nearest neighbor classifiers exploit notions of proximity and spatial regularity. Modern manifold learning algorithms such as Laplacian Eigenmaps and diffusion maps attempt to find hidden geometric structures within data. Without a question, geometry of the features play an important role in both supervised and unsupervised learning. 

While it was thought that the ideal geometries or configuration of features should vary by dataset, data type, number of dimensions, and feature transform, there is growing empirical evidence that there are universal and optimal configurations, and that appropriately trained neural networks yield such configurations \cite{papyan2020prevalence}. More precisely, that paper coined the term ``neural collapse" to refer to a phenomenon where, informally speaking, after completion of the training process, the neural network maps all data points from a single class to the same point in feature space, and that these points after recentering to their mean form the vertices of a (possibly rotated and scaled) simplex ETF (equiangular tight frame). There are several theoretical explanations for why neural collapse occurs -- typically arguing that a simplex ETF frame minimizes a certain functional under constraints on the features \cite{papyan2020prevalence,E2020emergence,mixon2020neural} -- but there are many open questions as to why neural networks do this to begin with and what the generalization advantages are.

\begin{figure*}[t]
    \centering
    \includegraphics[width=0.24\textwidth]{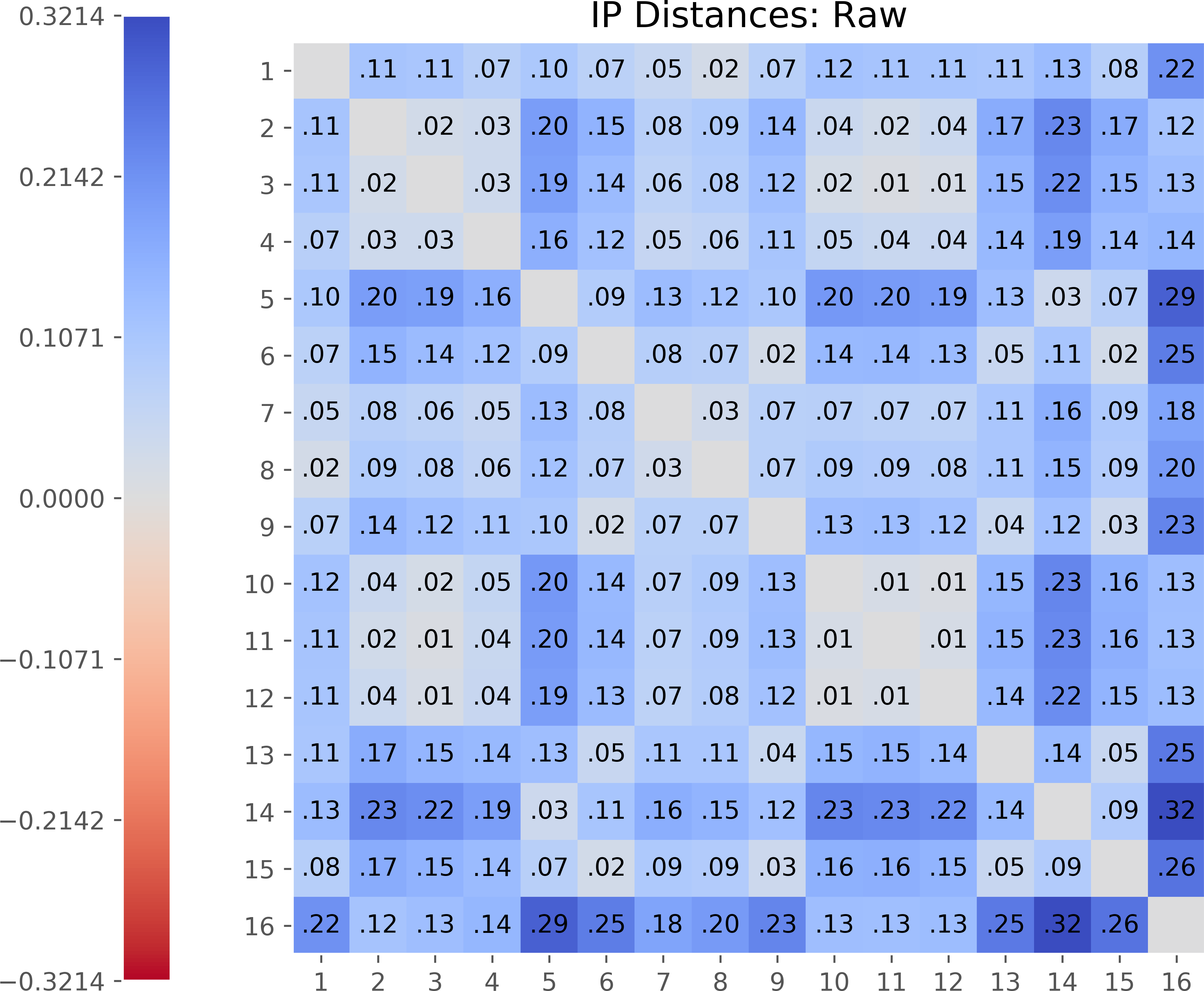}
    \includegraphics[width=0.24\textwidth]{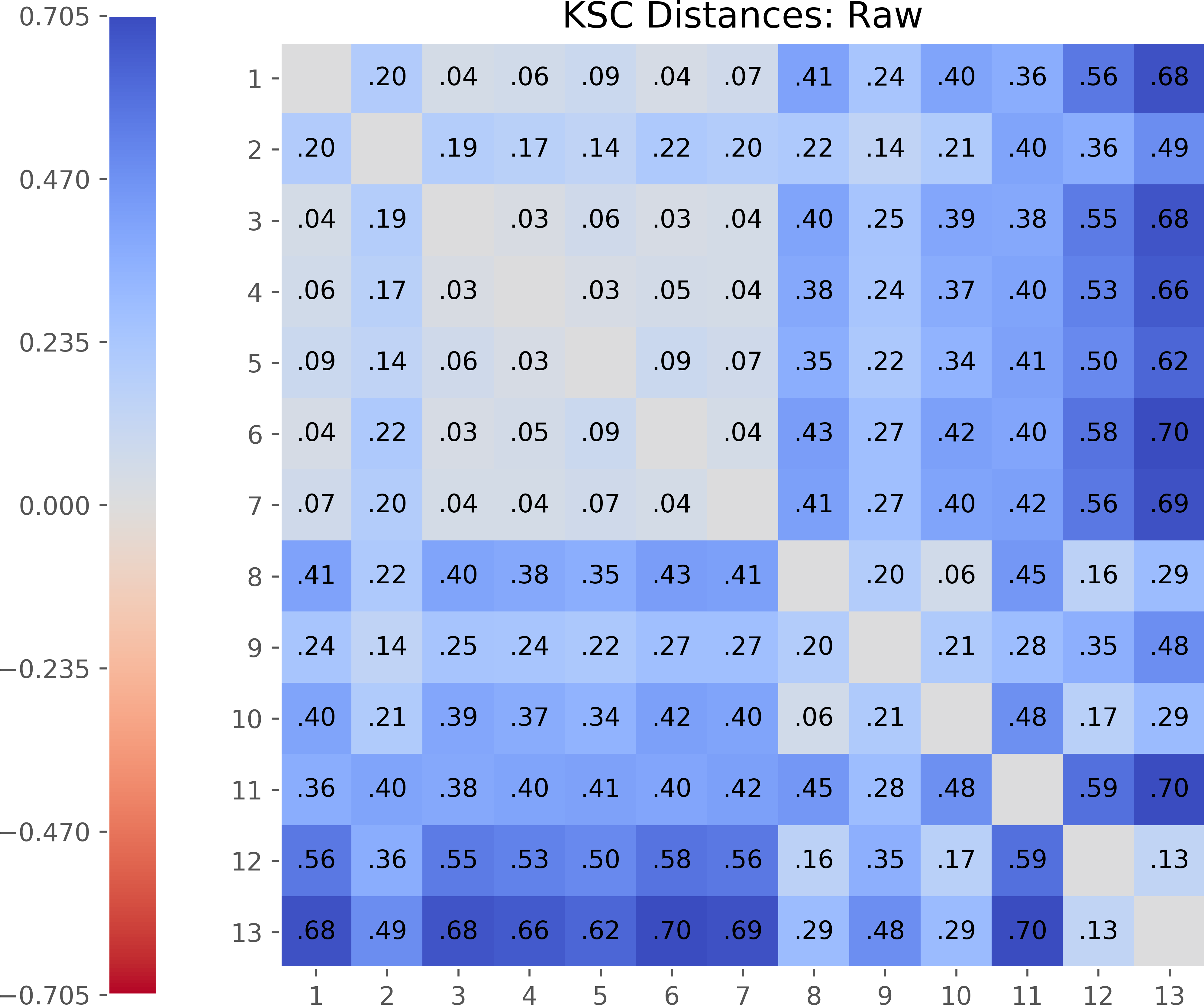}
    \includegraphics[width=0.24\textwidth]{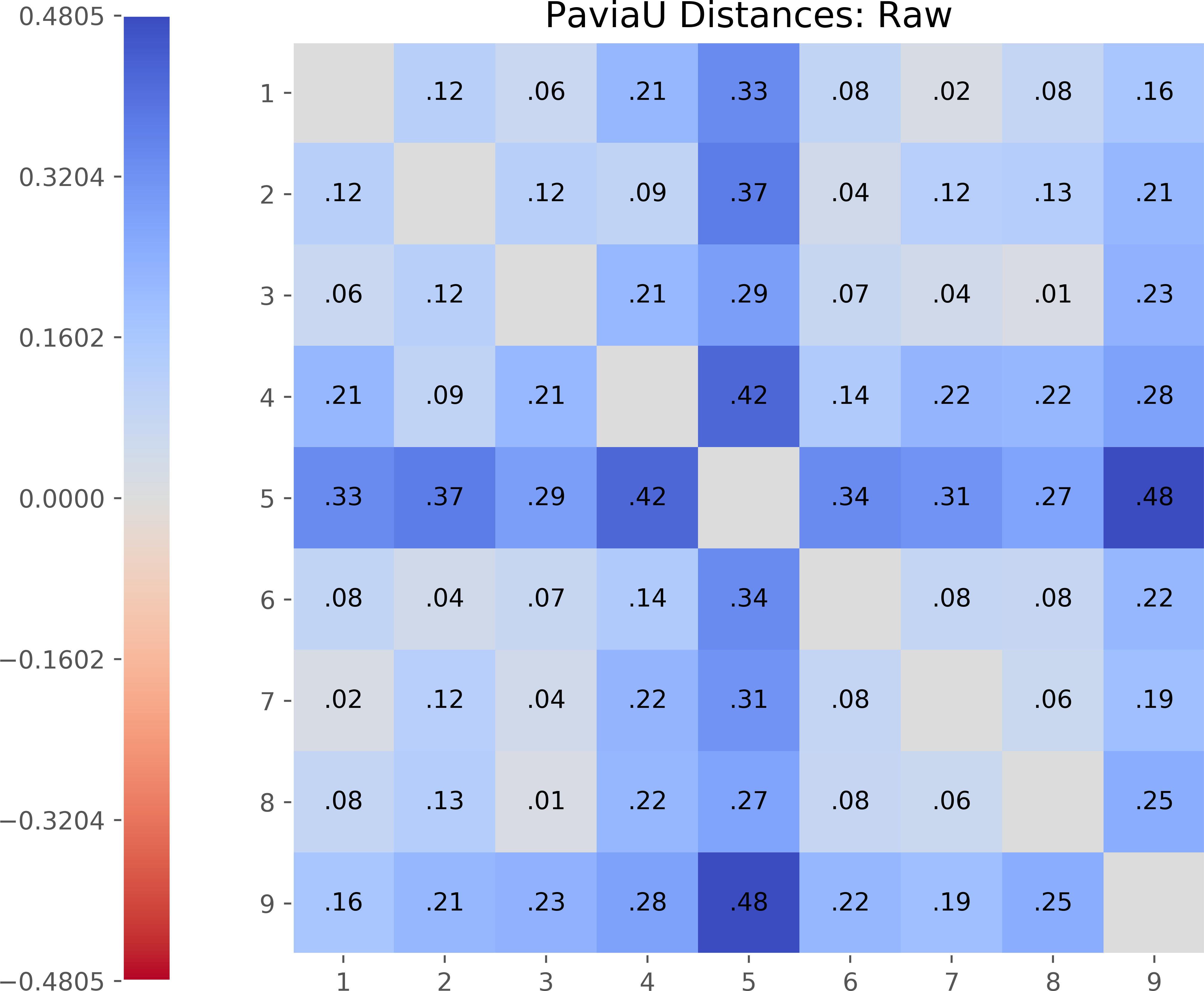}
    \includegraphics[width=0.24\textwidth]{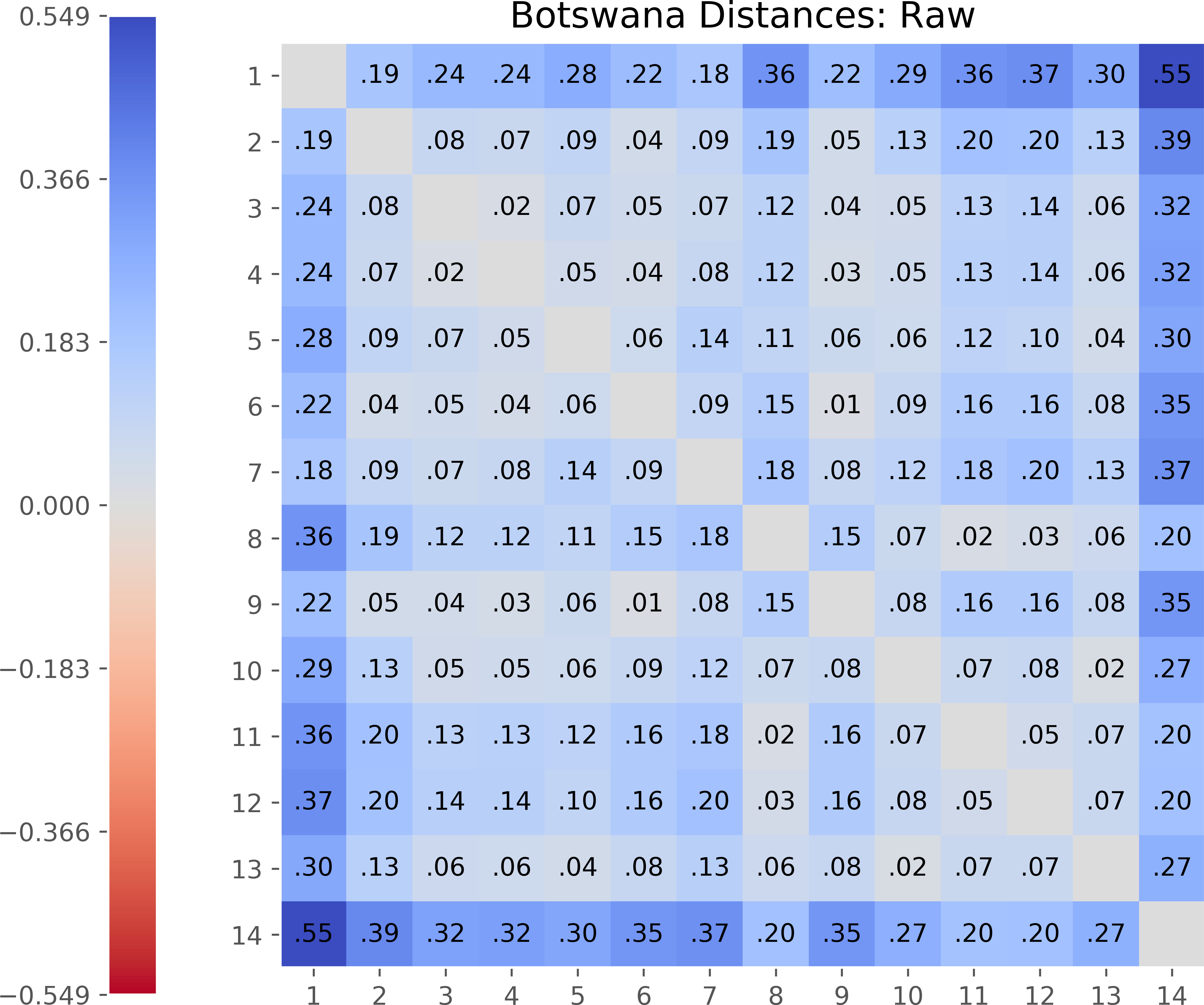} \\
    \includegraphics[width=0.24\textwidth]{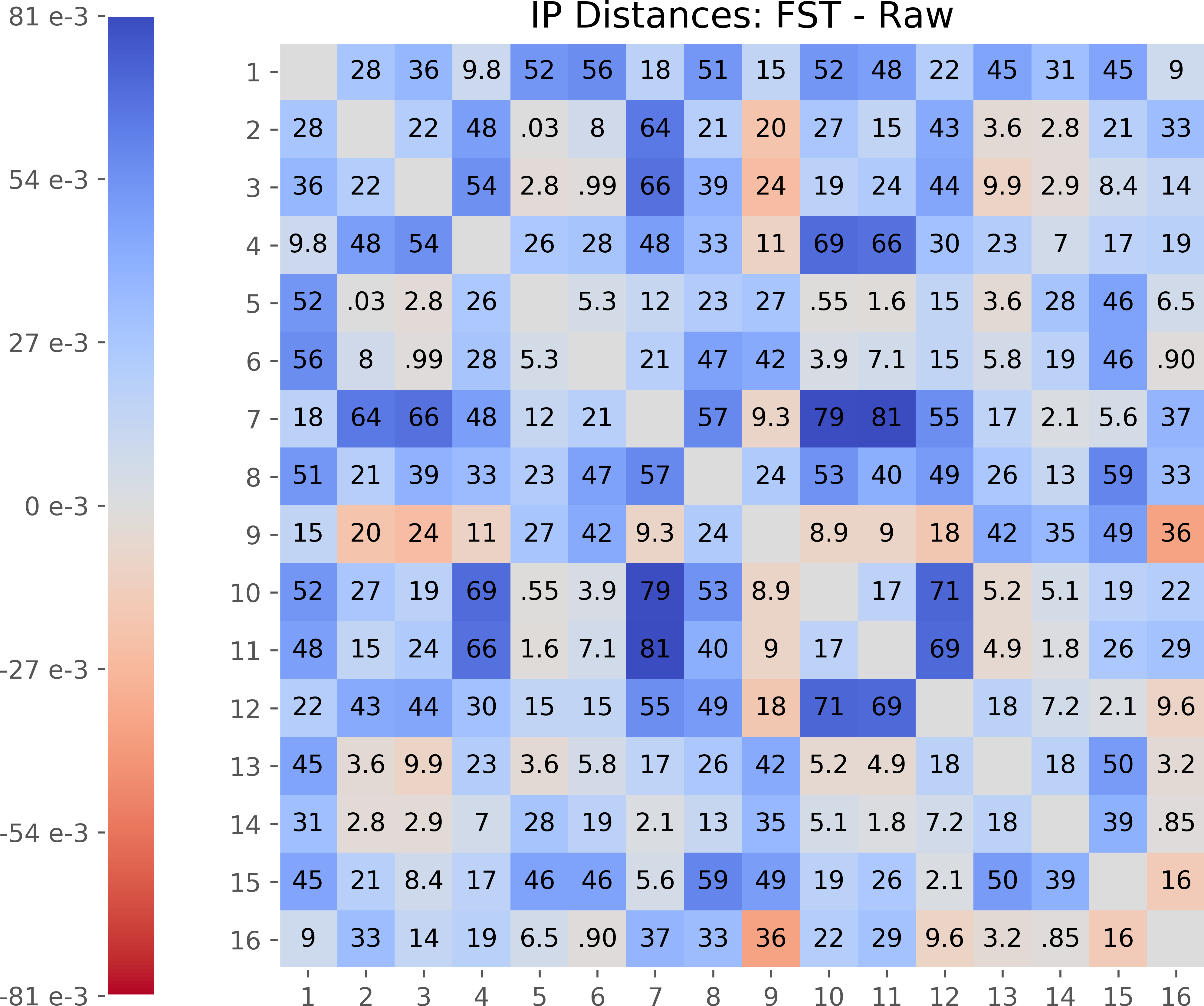}
    \includegraphics[width=0.24\textwidth]{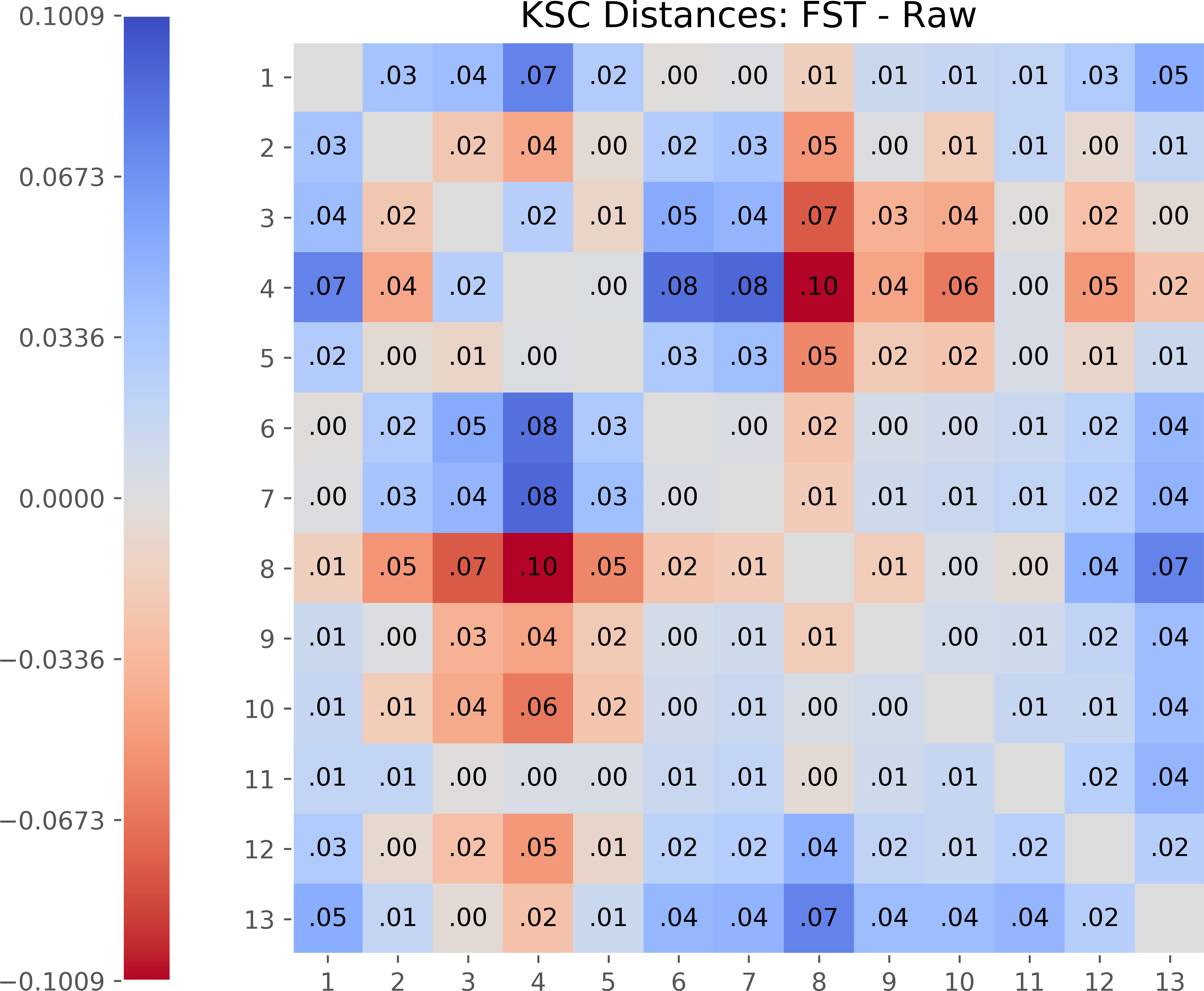}
    \includegraphics[width=0.24\textwidth]{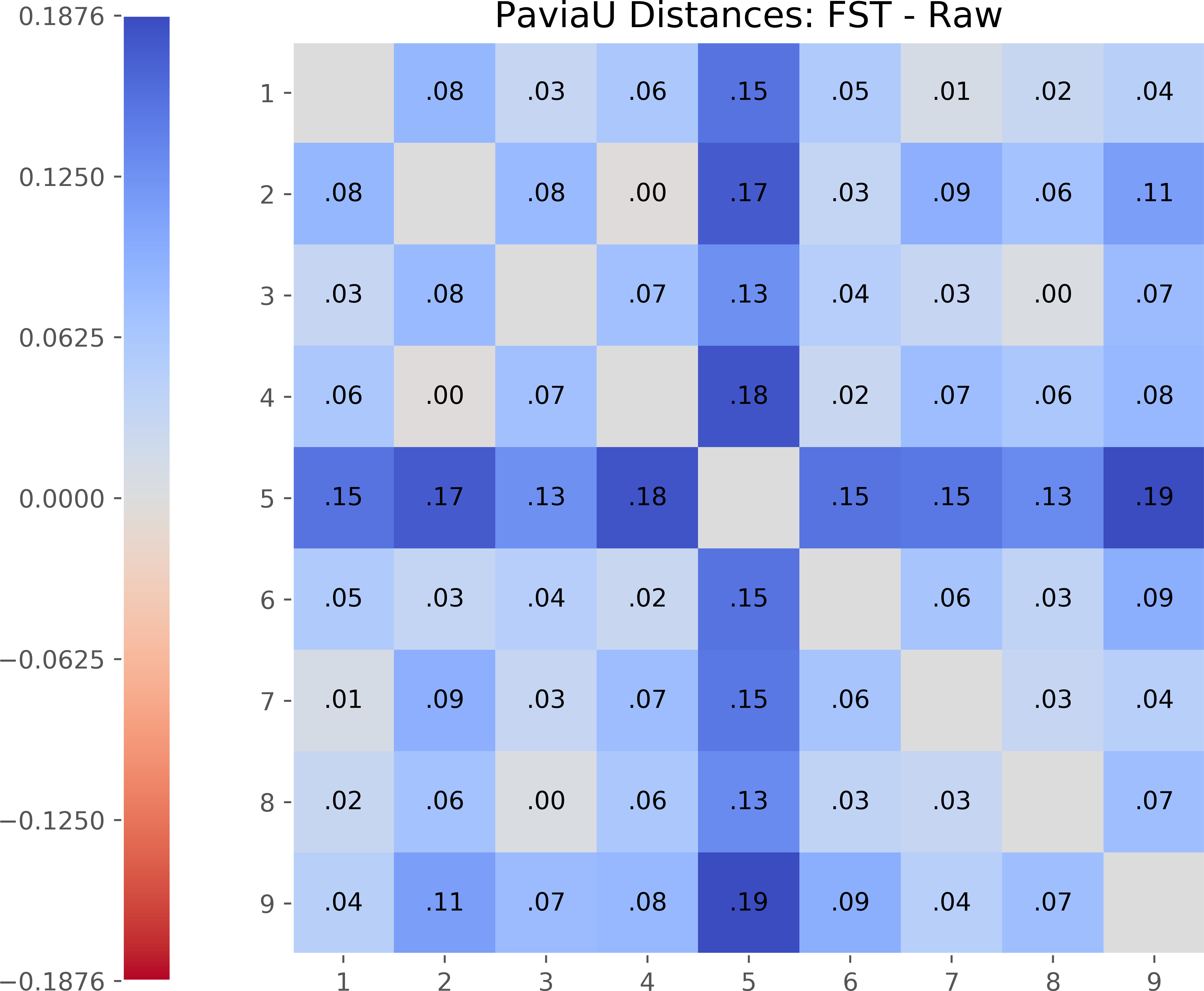}
    \includegraphics[width=0.24\textwidth]{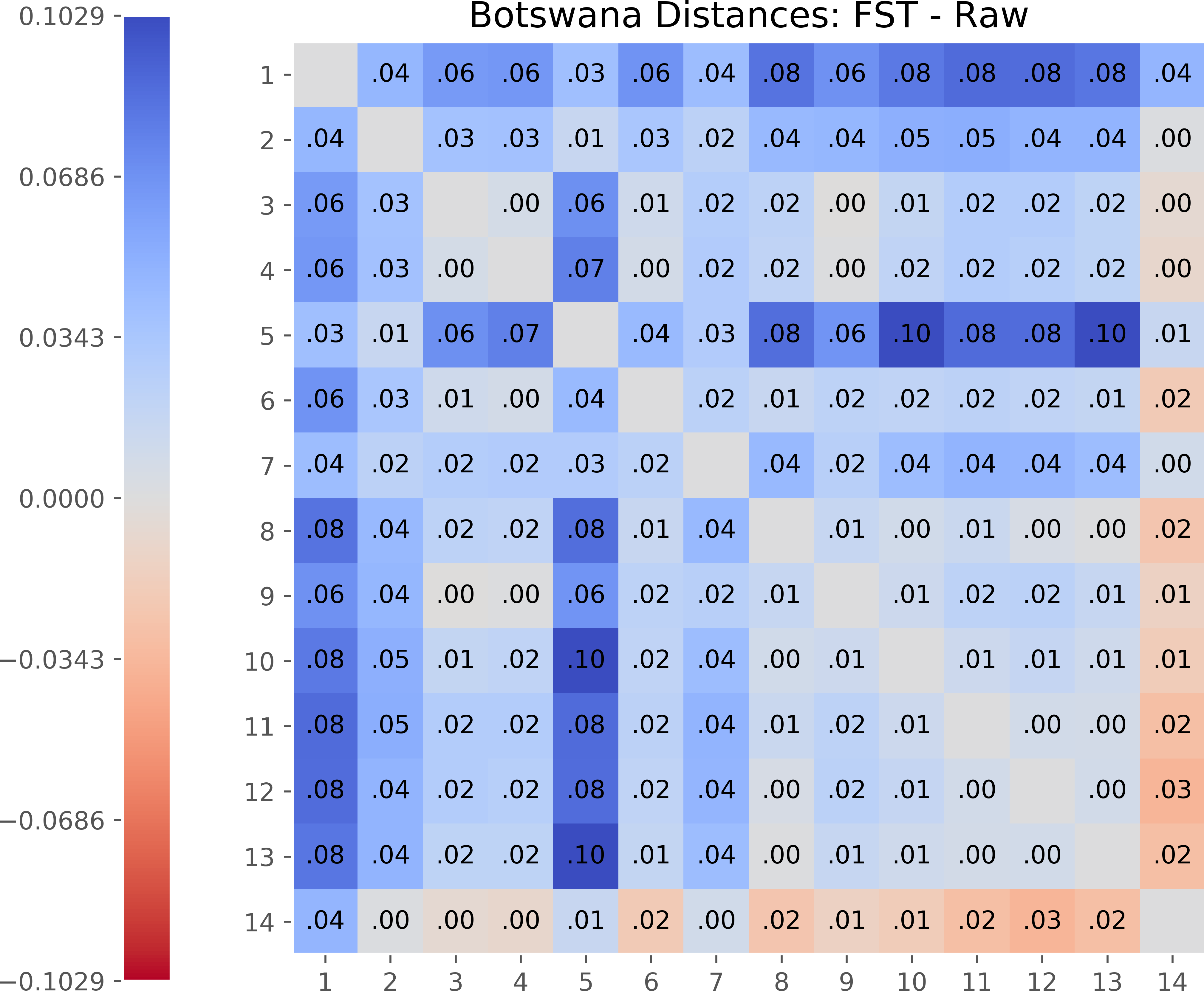} \\
    \includegraphics[width=0.24\textwidth]{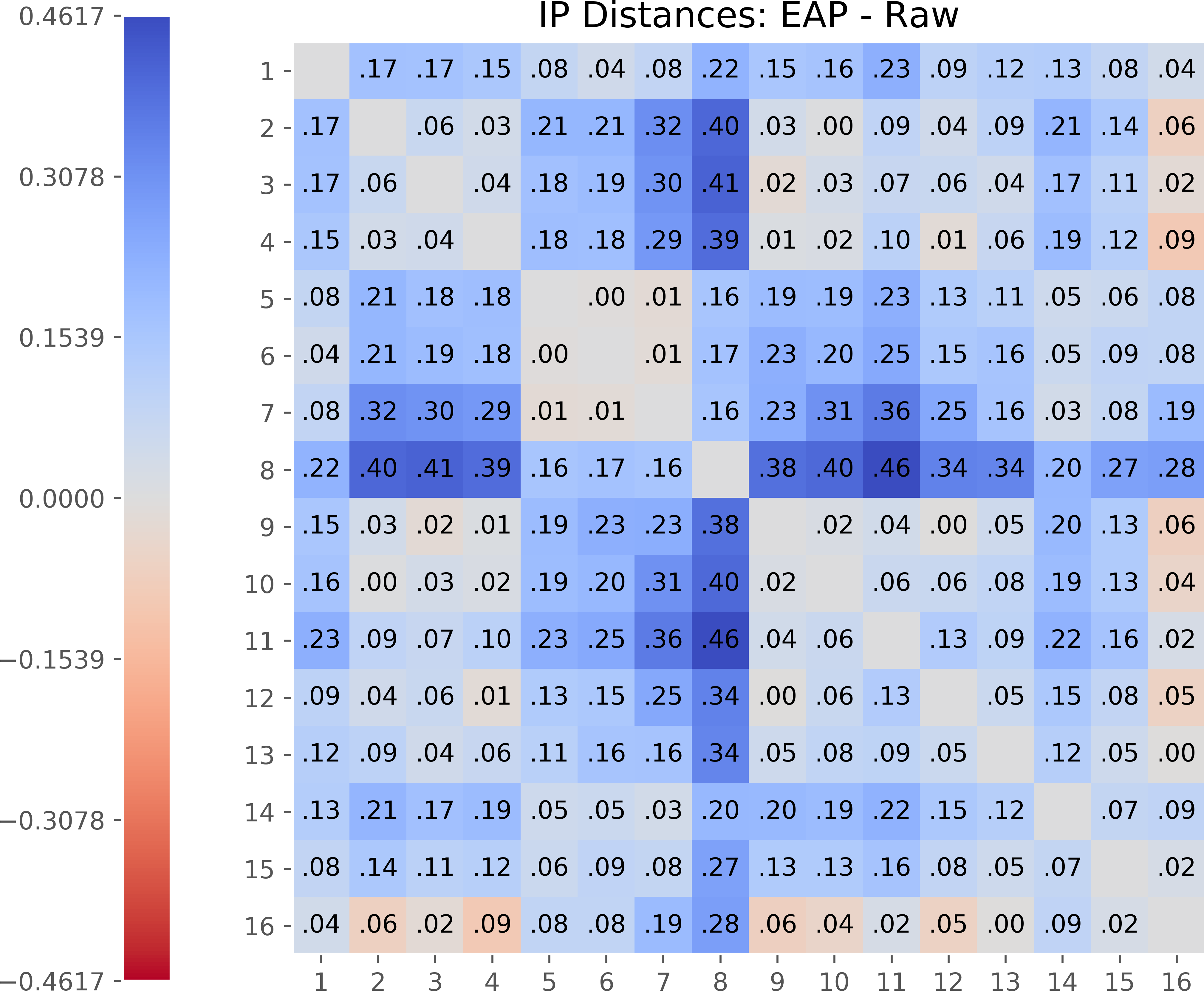}
    \includegraphics[width=0.24\textwidth]{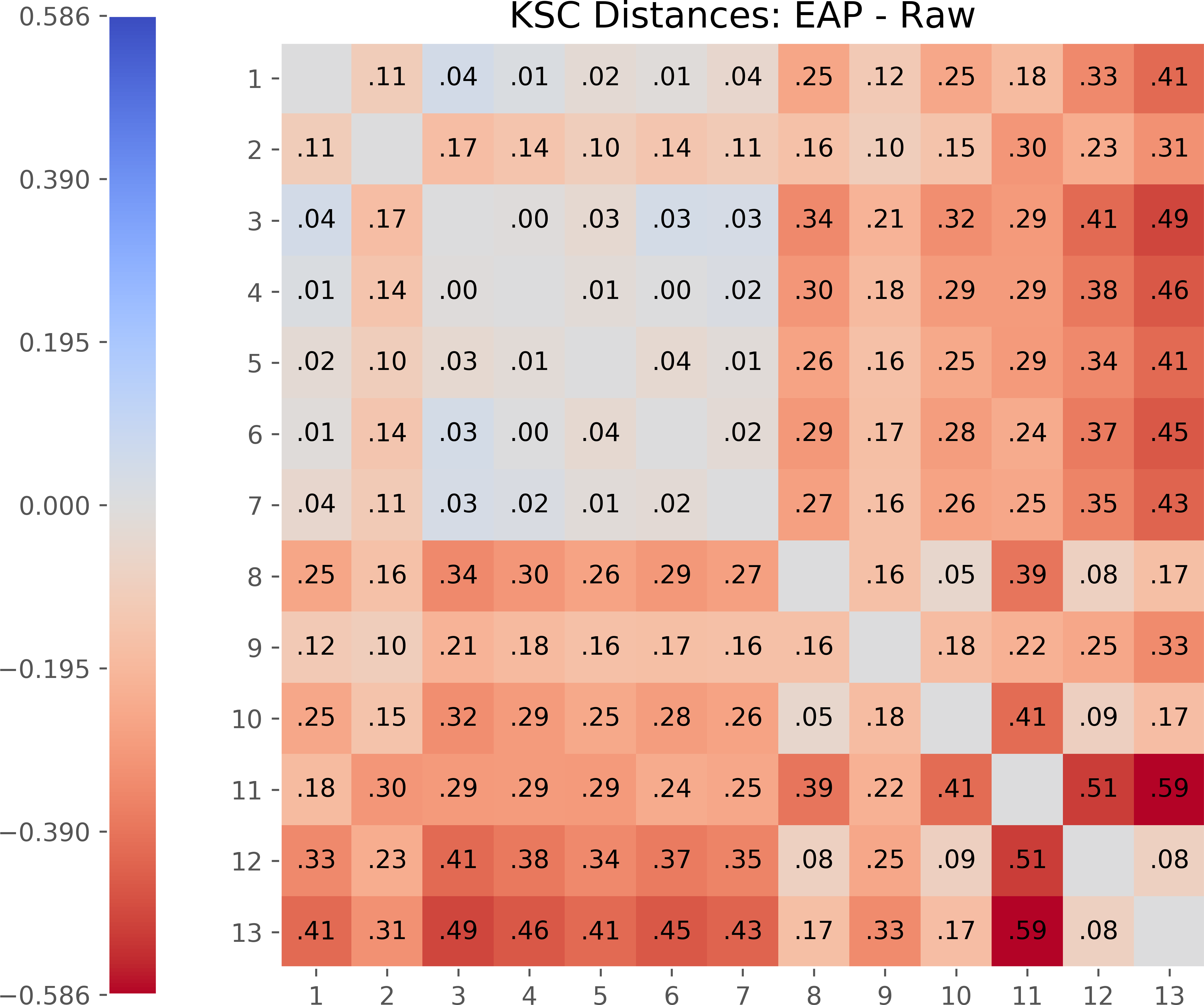}
    \includegraphics[width=0.24\textwidth]{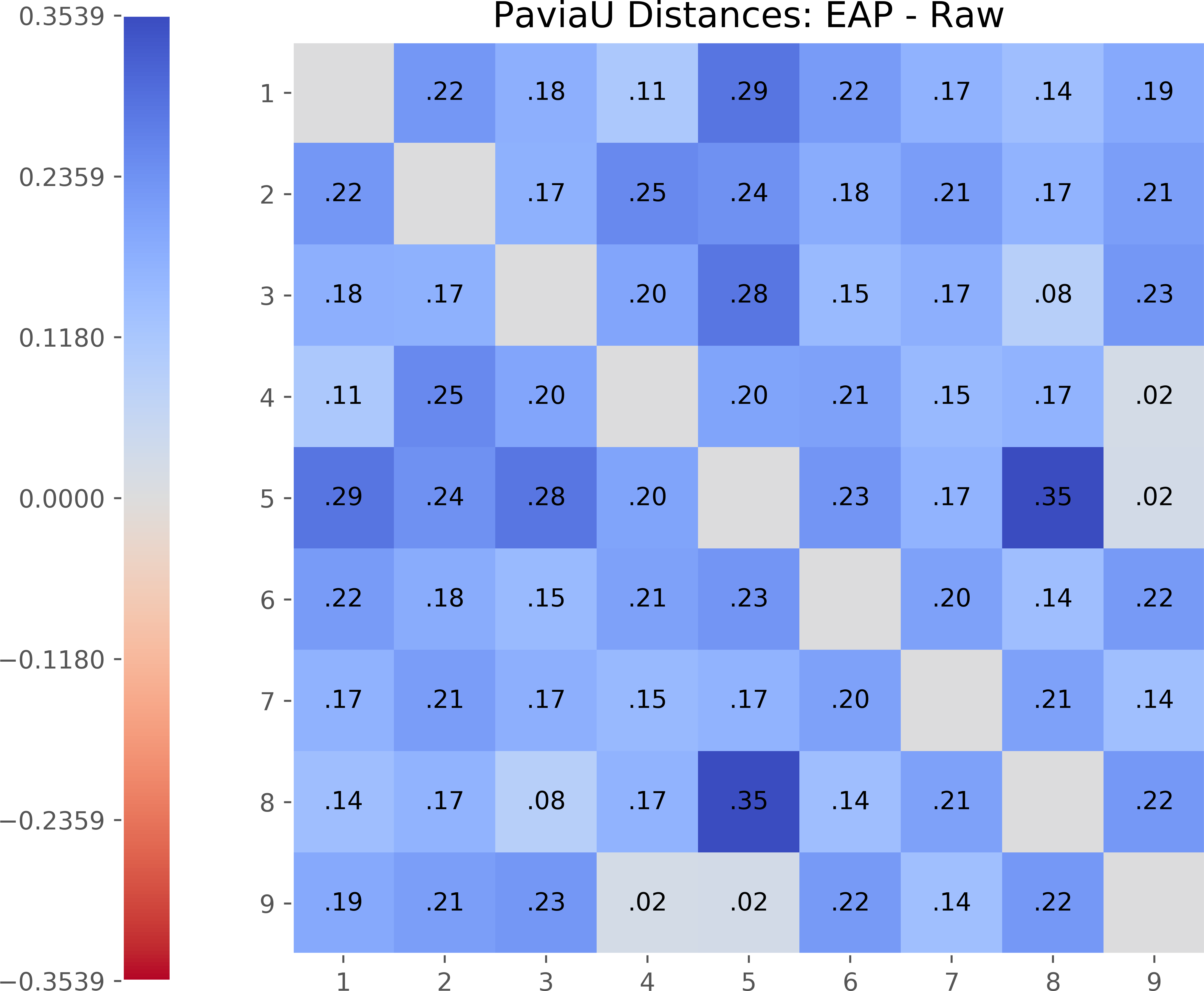}
    \includegraphics[width=0.24\textwidth]{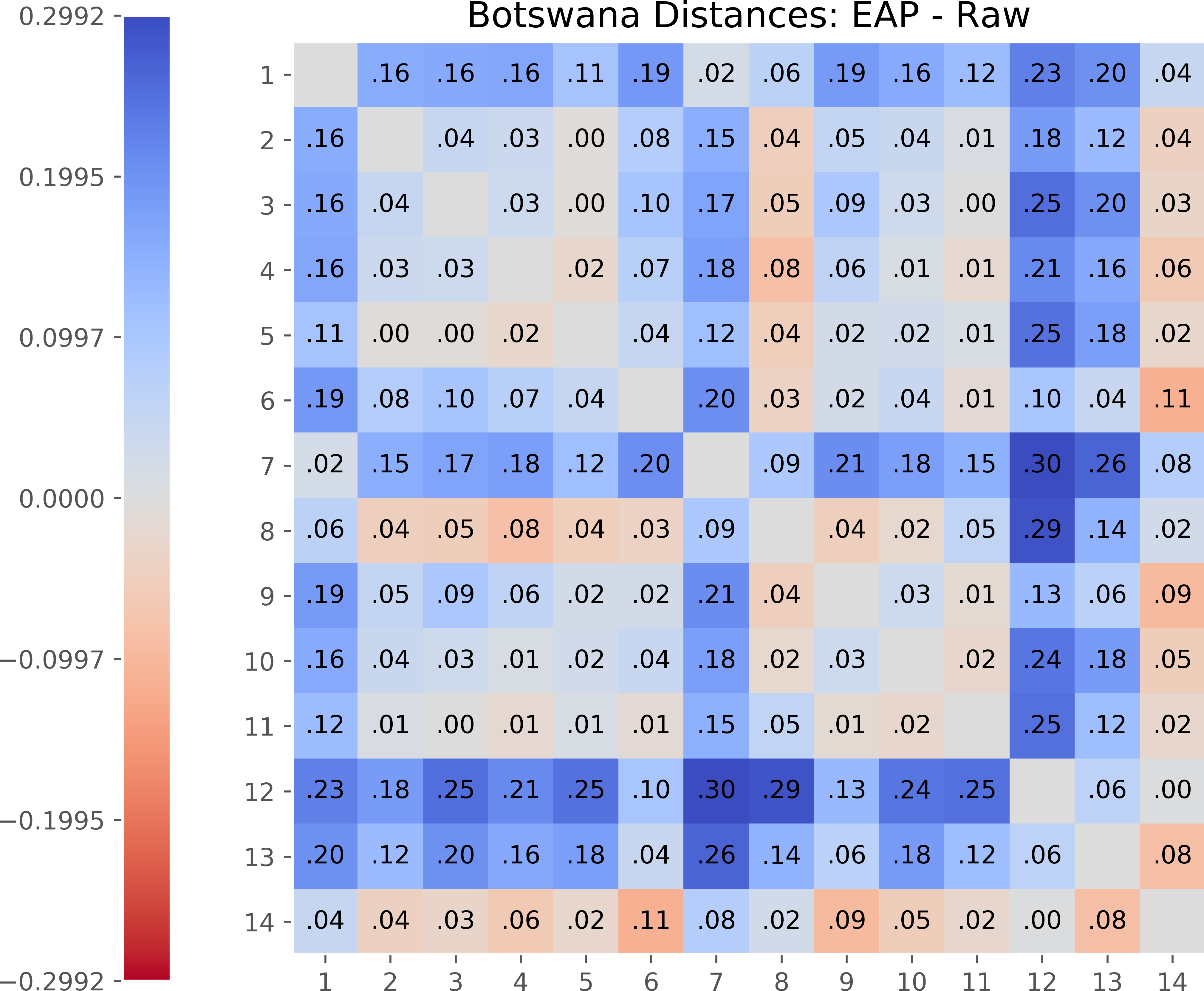} 
    \caption{Top row: class mean distances for raw features. Middle row: 3DFST minus raw. Bottom row: EAP minus raw.}
    \label{fig:dist}
\end{figure*}

Inspired by recent innovations, we examine the geometry of the features induced by two state-of-the-art methods for HSI classification. 
\vspace{-.5em}
\begin{itemize}\itemsep -0.5em
    \item 
    3DFST (three-dimensional Fourier scattering transform) is a feature extractor formed by cascading the complex modulus with convolutions against functions generated by time-frequency shifts. This is a data-independent transform that combines the strengths of spectral decomposition with hierarchical methods \cite{czaja2019analysis,czaja2020rotationally}, and was used for HSI classification \cite{kavalerov2020,czaja2018scattering} and radioactive anomaly detection \cite{peterson2019experiments}.
    \item
    EAP (deep learning with extended attribute profile) \cite{aptoula2016deep,eap_code} projects the HSI spectra to several PCA components preserving most of the spectral energy and then creates attribute profiles (APs) for the 2D images corresponding to each component. The APs connect similar 2D structures through the morphological operations of erosion and dilation and create a more homogeneous version of the original 2D principle component image. Finally the cube of EAPs is passed to a 5 layer Neural Network which performs the classification.
\end{itemize}
Due to space constraints, we refer the reader to above references for further explanations. 

We examine the features generated by both methods on four standard hyperspectral datasets: Indian pines, Kennedy Space Center, Botswana, and Pavia University. We emphasize that the purpose of this proceeding is to examine the geometries of the features, not to compare their classification performances. For the later, we refer the reader to \cite{kavalerov2020}.

\vskip-0.2cm
\section{Methodology}
\label{sec:methodology}
\vskip-0.2cm
For each hyperspectral dataset and choice of features (raw, 3DFST, or EAP), we have access to a collection of vectors $\{\x_k\}_{k=1}^n$ in $\R^p$ with associated labels $\{y_k\}_{k=1}^n$. It is worth mentioning that the focus of this paper is not on classification performance -- there is no training or test split in this paper. Here $n$ is the total number of labeled points in the HSI. 

The dimension $p$ of the feature domain varies by dataset and algorithm, so in order to have a fair comparison, we normalize each $\x_k$ by dividing by the maximum Eulidean norm over all the features. Doing so, we can assume each $\x_k$ is contained within the unit Euclidean ball of their respective domains. This has no impact on the results we present below, since we will examine quantities that are invariant to global scaling. Throughout, we only consider the Euclidean distance $\|\cdot\|$ between points. 
\par 
We briefly summarize the parameters of the methods we use. All code for our feature extraction methods is open sourced at \url{https://github.com/ilyakava/pyfst}.

3D FST performs 3 stages of 3D filters, interspersed with downsampling and non linearities. For these hyperparamters we follow \cite{kavalerov2020} which found the best sizes of 3D filters to use for HSI 3D Scattering via a grid search on small validation sets. These 3D filter sizes vary by dataset.
Classification is then performed with a linear svm with regularization parameter fixed to $C=1000$.

The EAP method performs PCA, finds morphological profiles, and then uses a neural network to classify. For the parameters of these we follow \cite{aptoula2016deep,eap_code}, which uses 4 PCA components and generates 9 morphological profiles for each component. Then, a square window of size $9$ in the spatial dimension is treated as a single input into the Neural Network, which generates a class prediction. This $9\times9\times 36$ cube is passed to a neural network containing 3 2D convolutional layers (with spatial sizes $5,3,3$ and channel lengths $48,96,96$) and 2 Fully Connected layers of size $1024$ before a softmax classifier. Thus the penultimate feature size before classification is always $1024$.
\vskip-0.2cm


\begin{figure*}[t]
    \centering
    \includegraphics[width=0.24\textwidth]{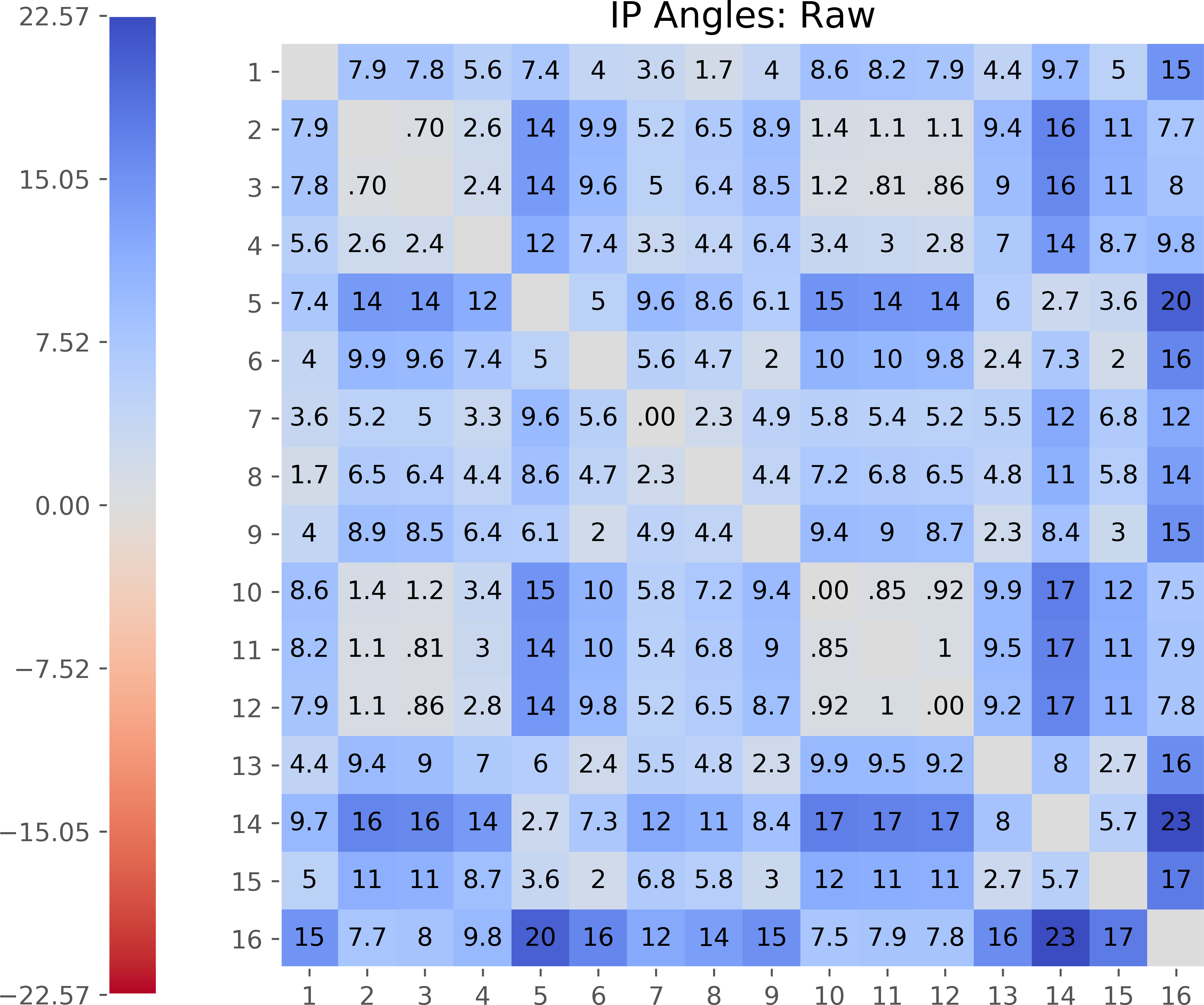}
    \includegraphics[width=0.24\textwidth]{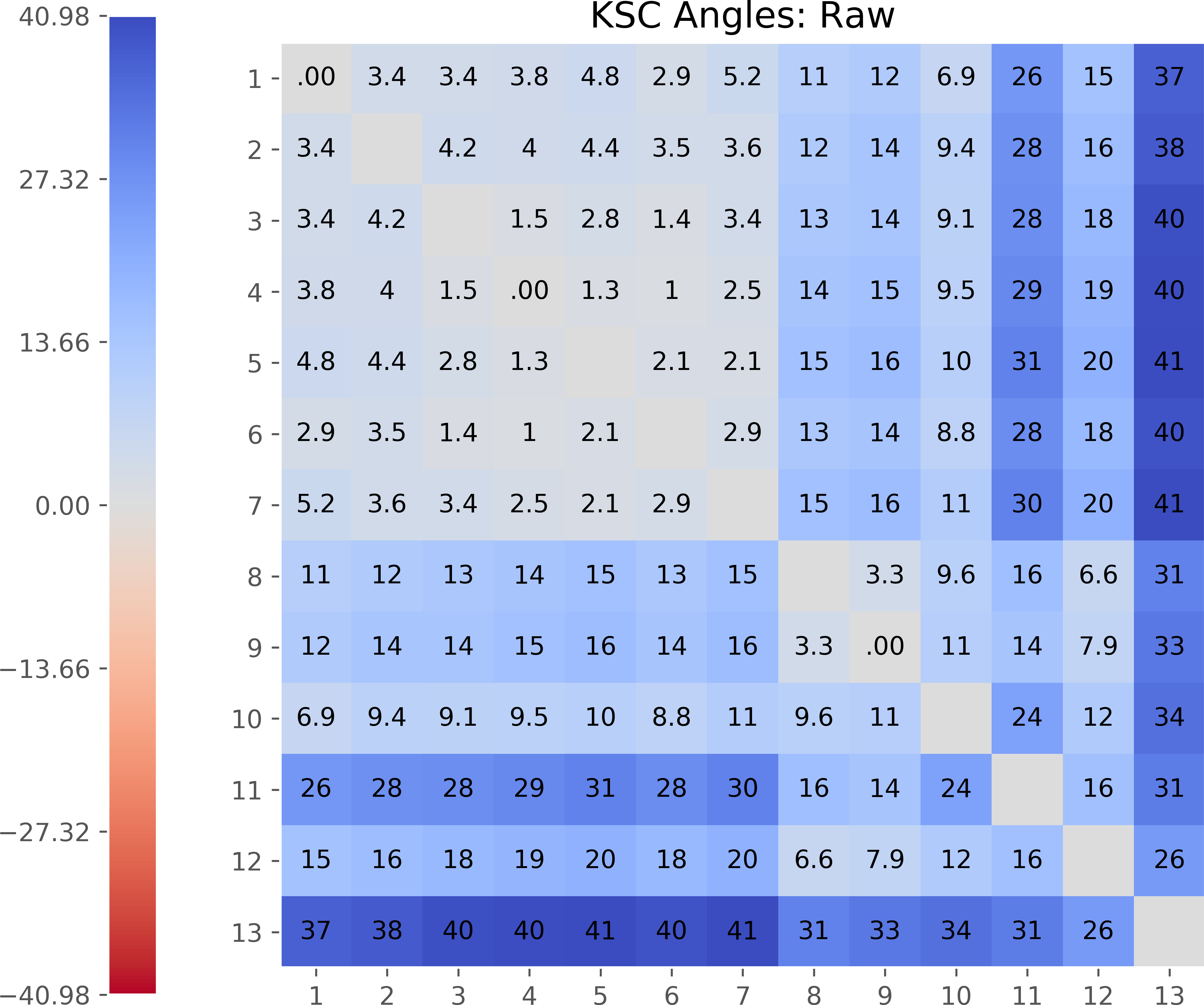}
    \includegraphics[width=0.24\textwidth]{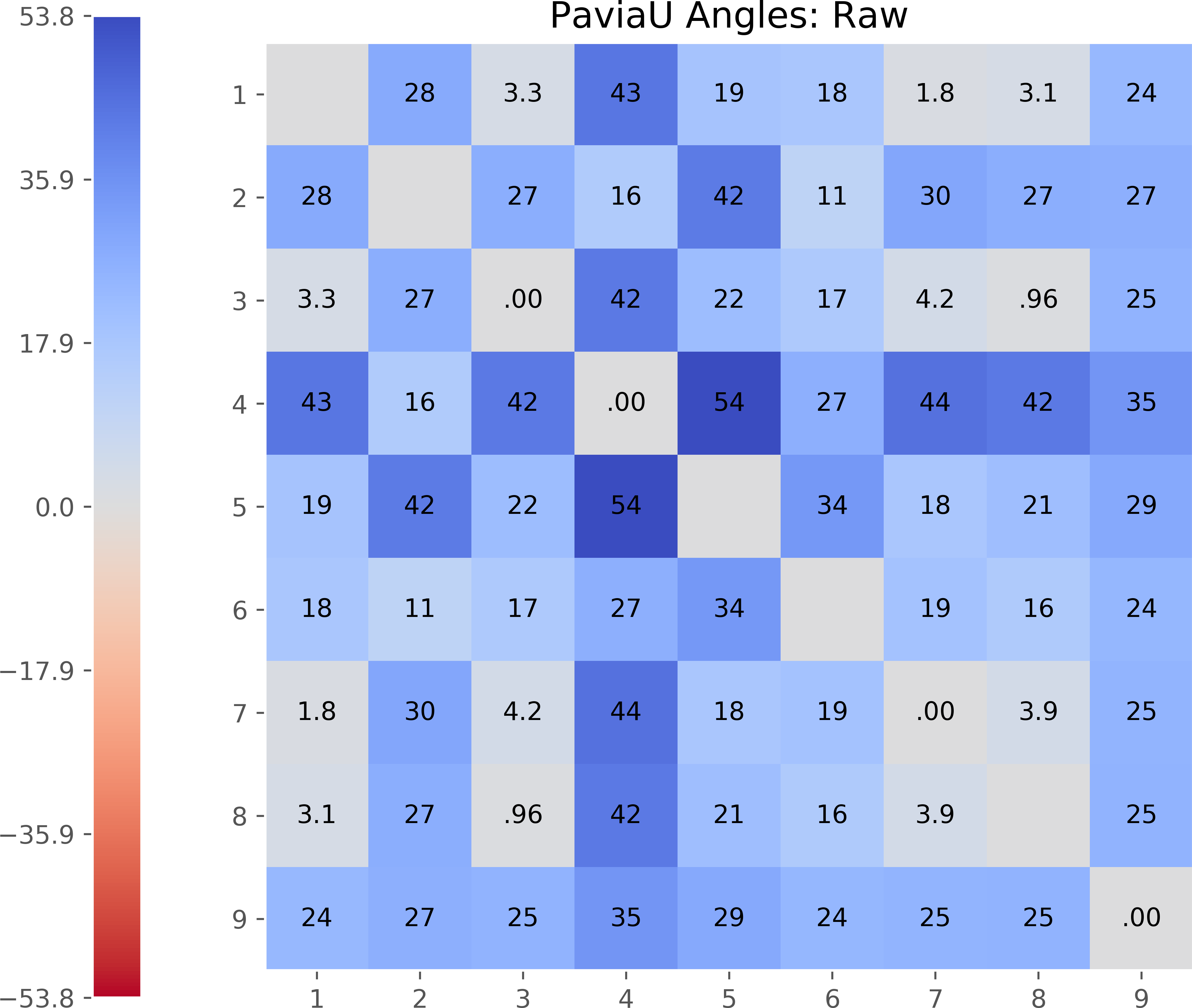}
    \includegraphics[width=0.24\textwidth]{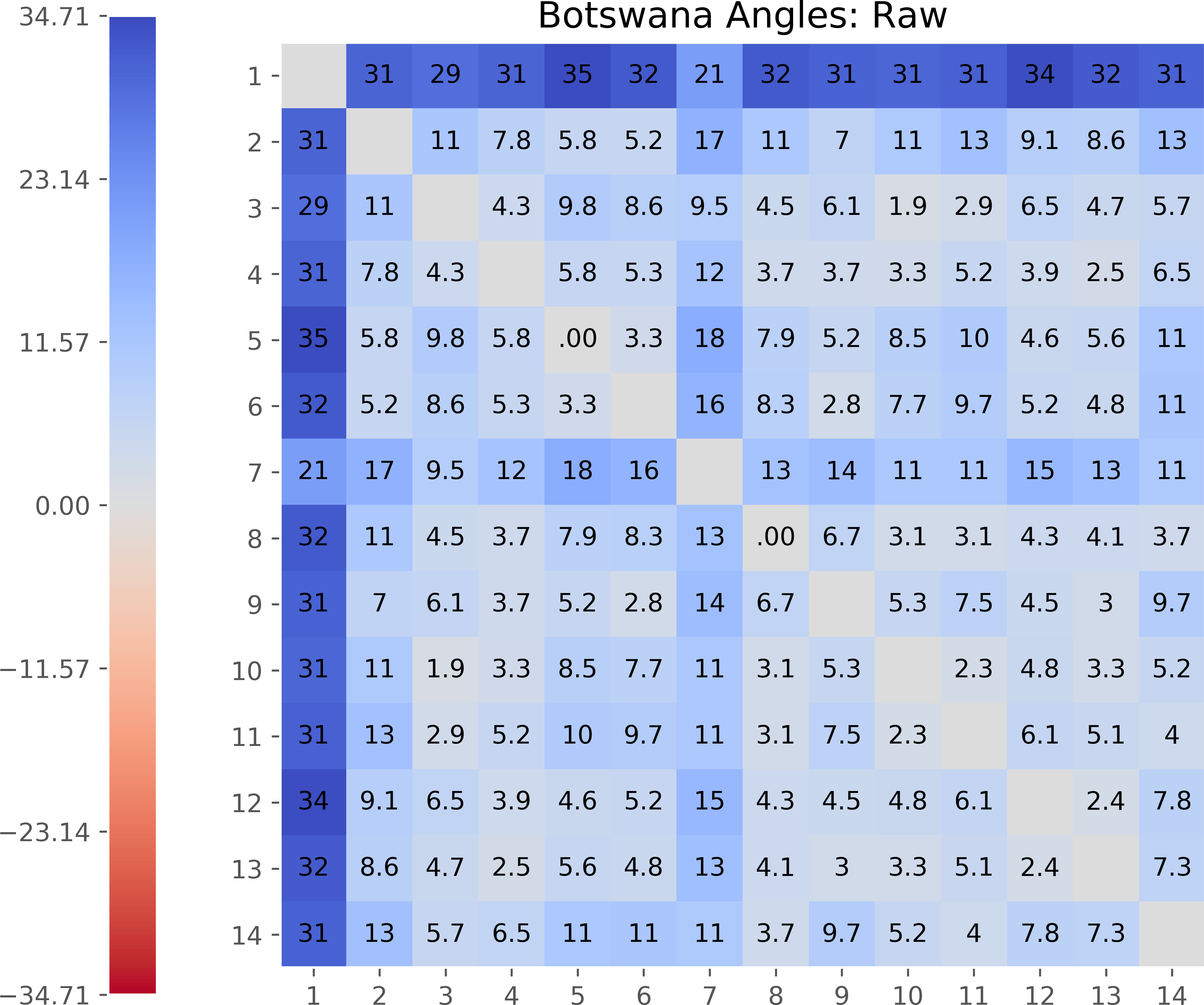} \\
    \includegraphics[width=0.24\textwidth]{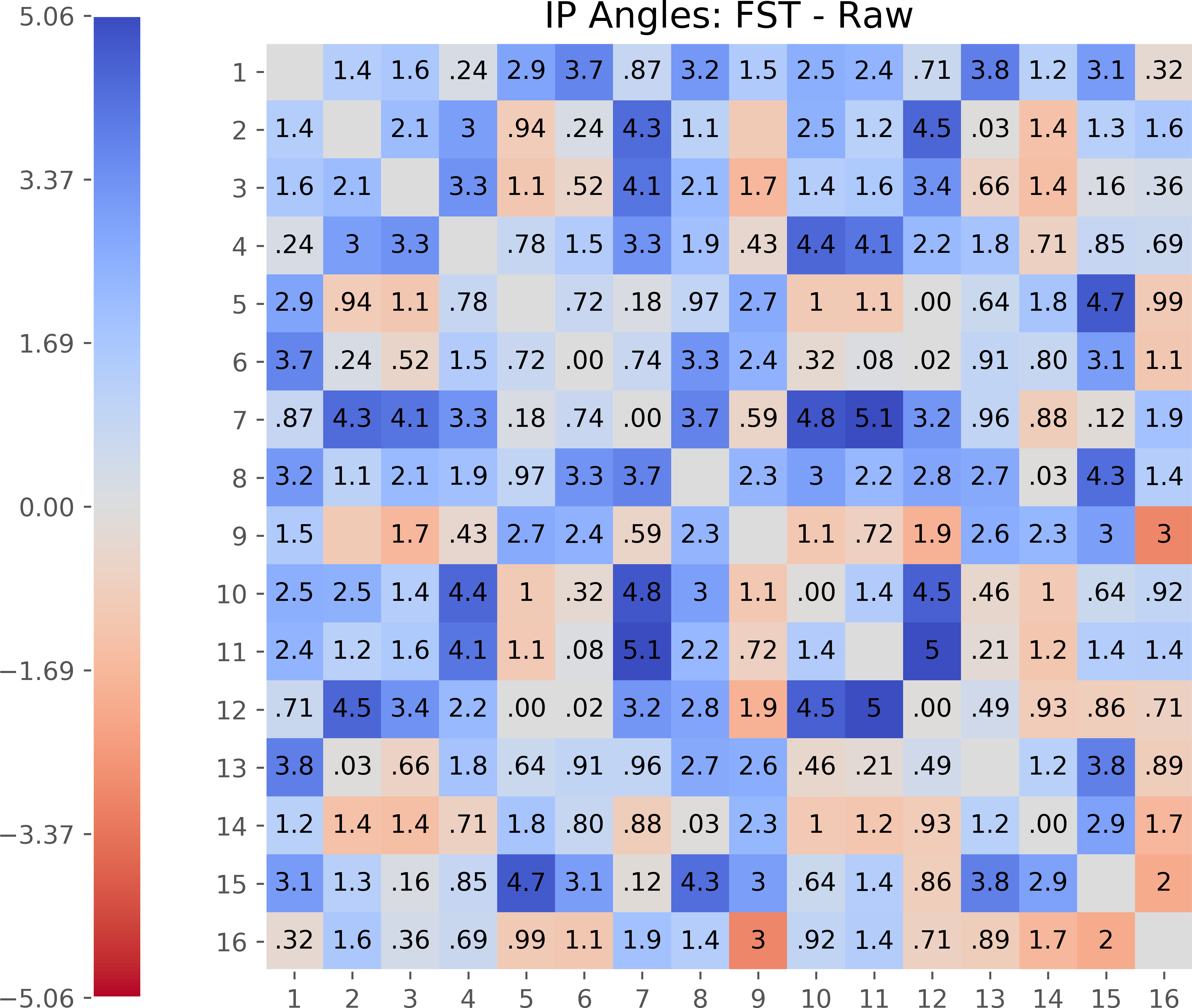}
    \includegraphics[width=0.24\textwidth]{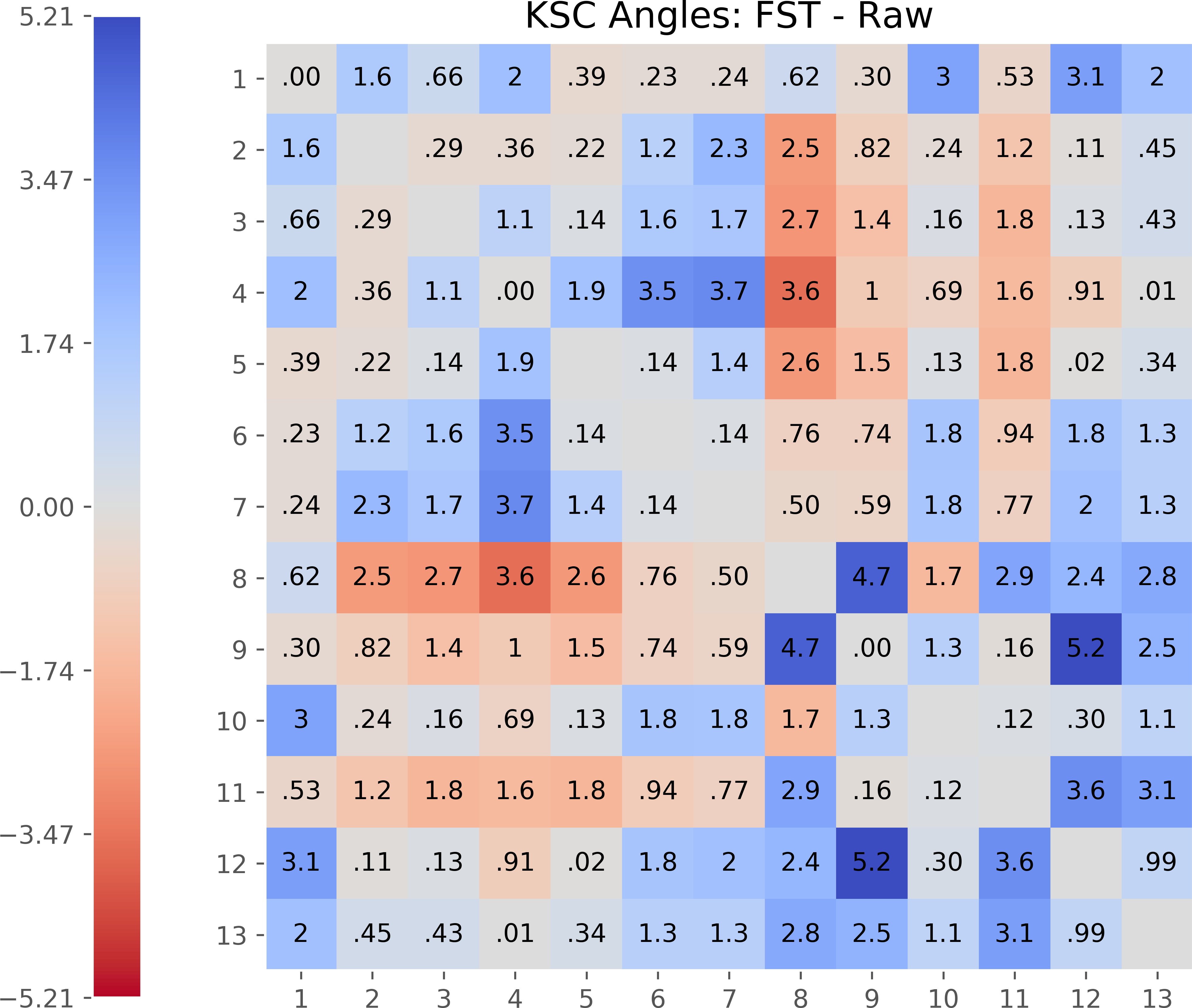}
    \includegraphics[width=0.24\textwidth]{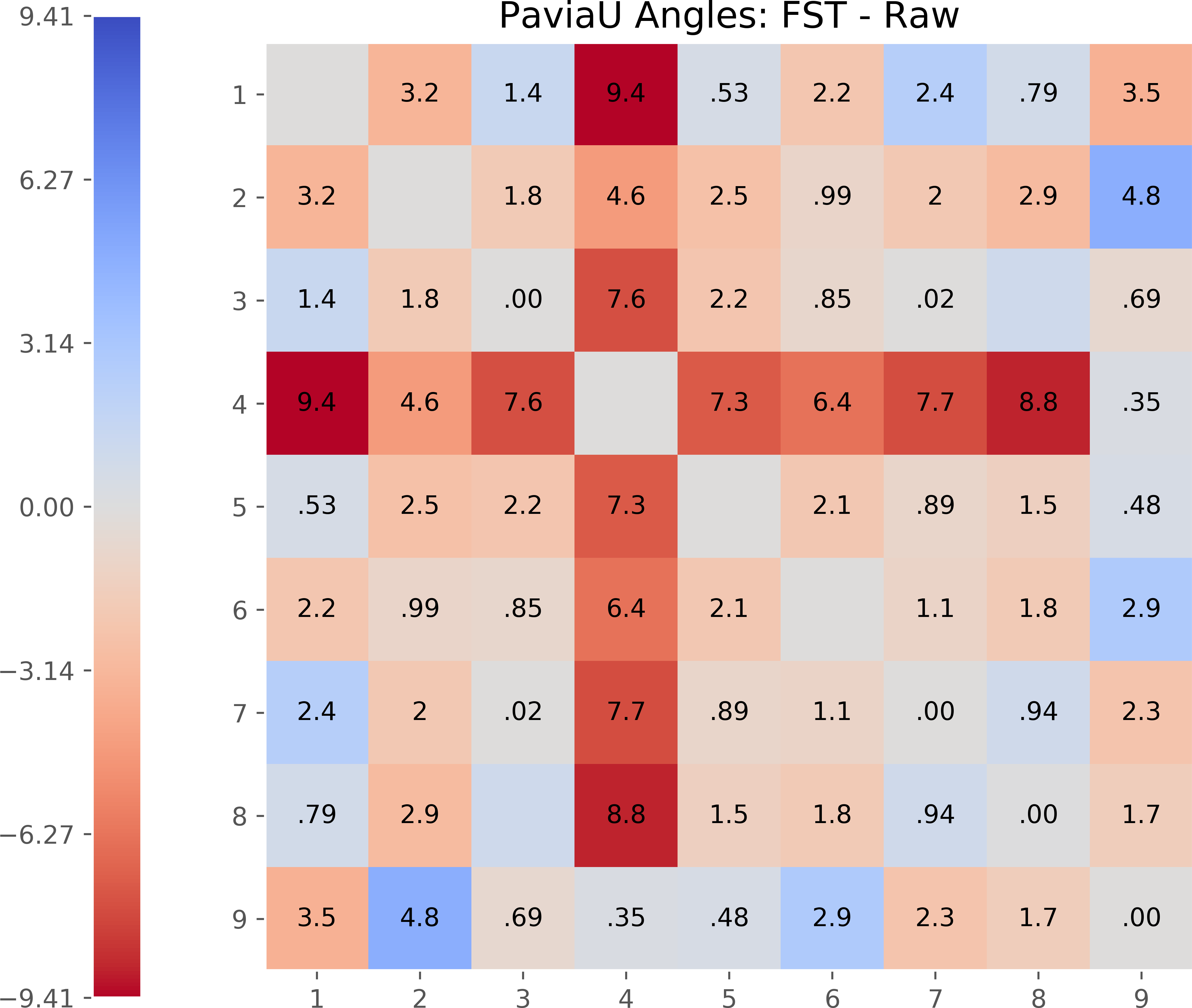}
    \includegraphics[width=0.24\textwidth]{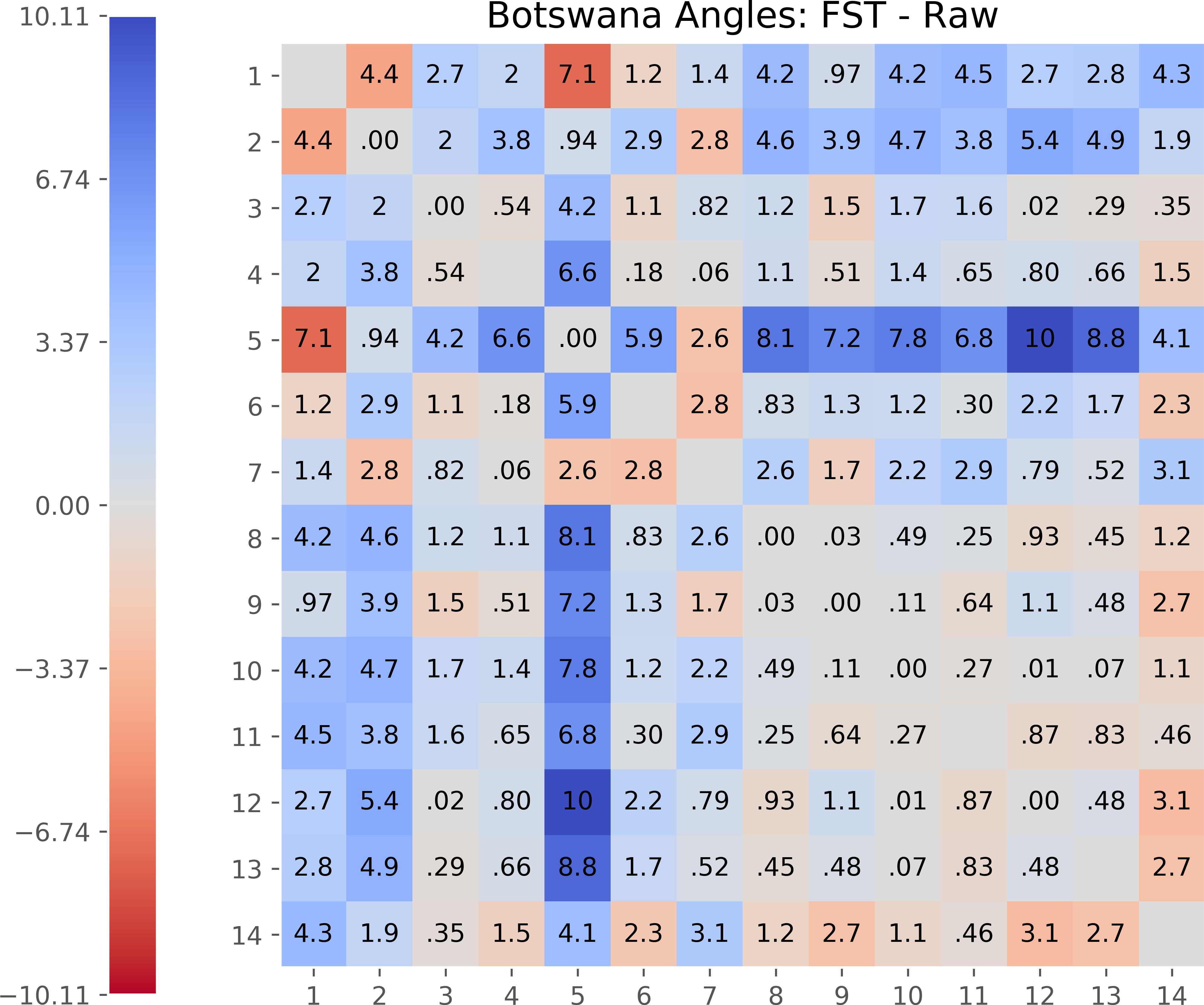} \\
    \includegraphics[width=0.24\textwidth]{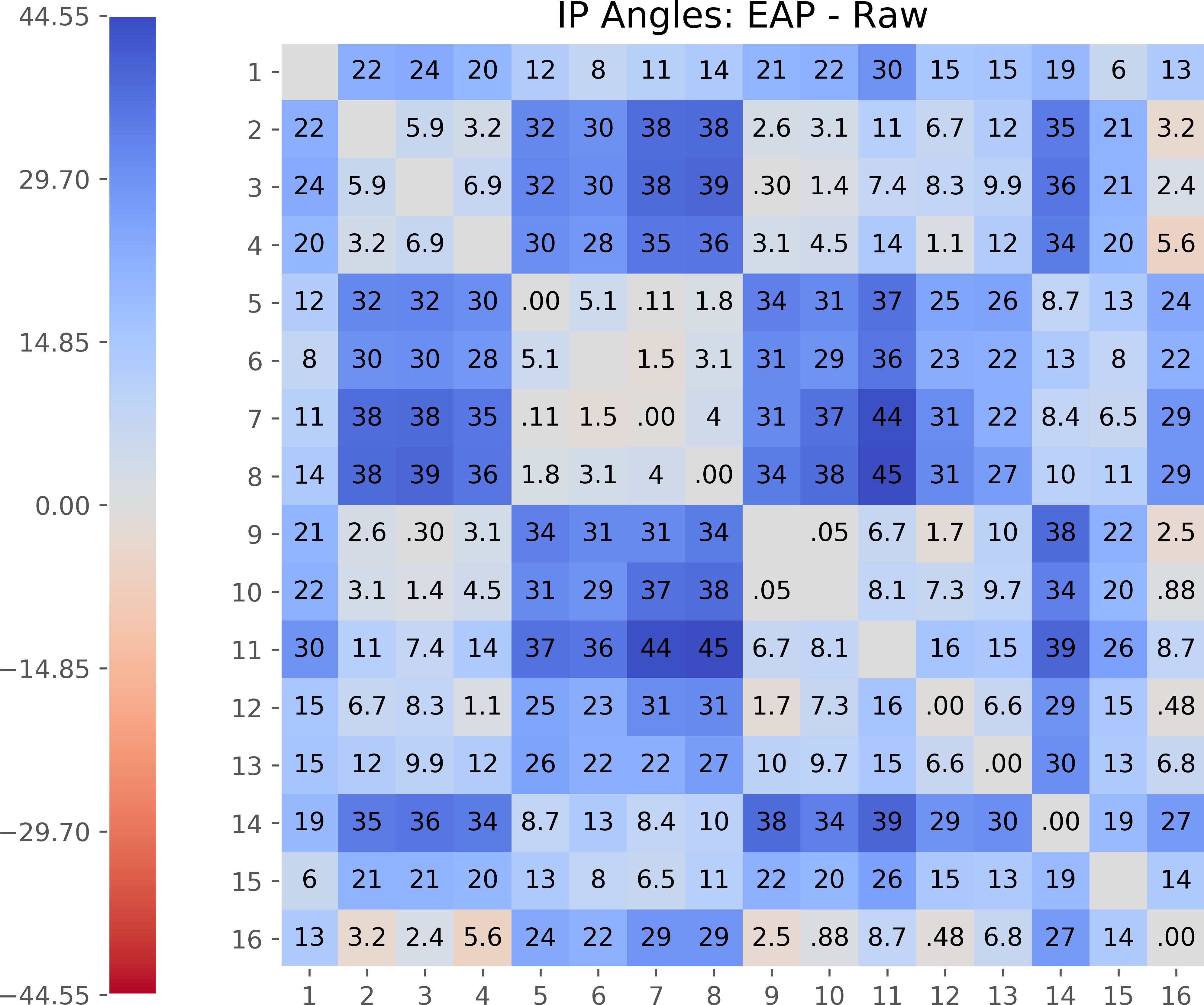}
    \includegraphics[width=0.24\textwidth]{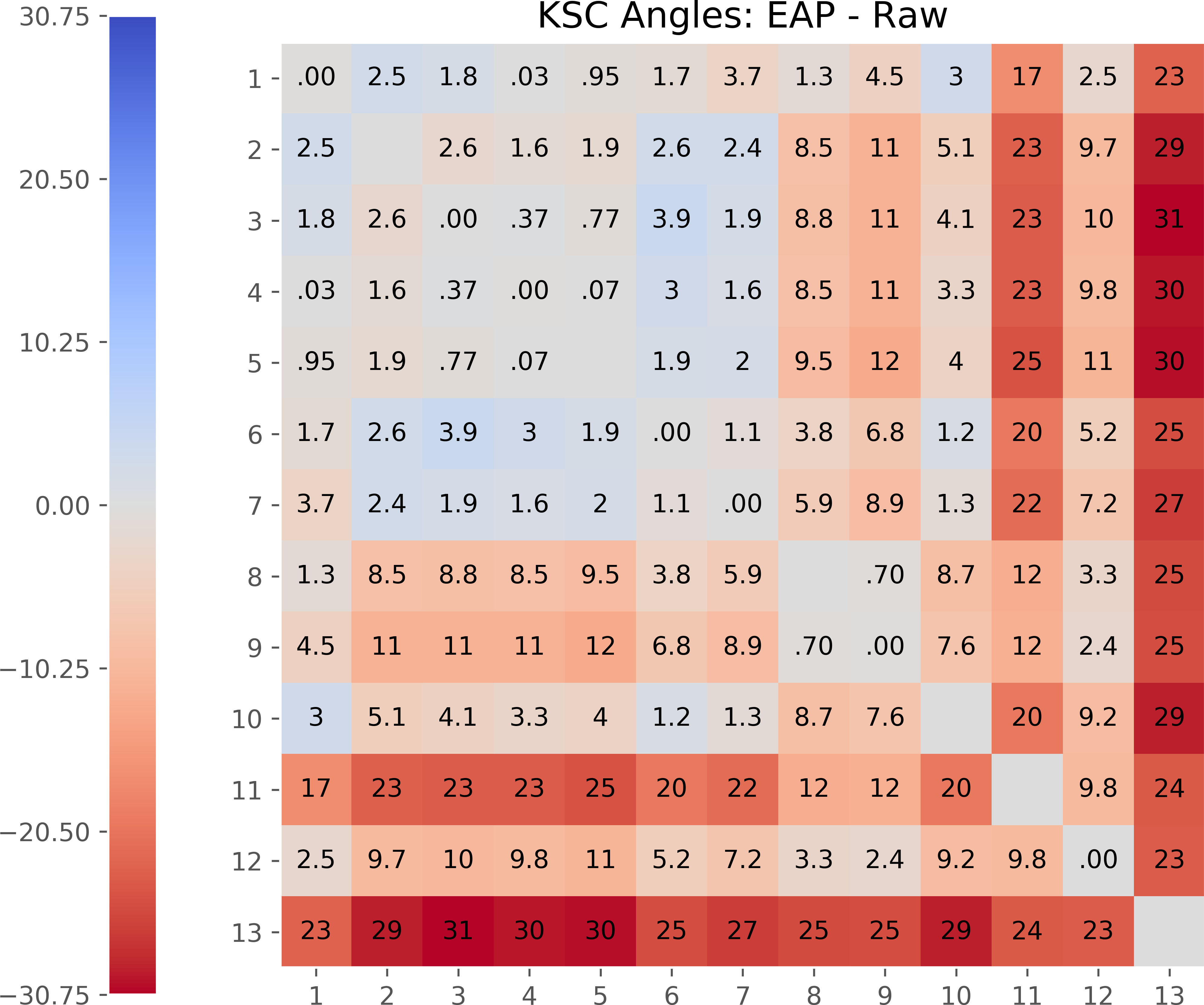}
    \includegraphics[width=0.24\textwidth]{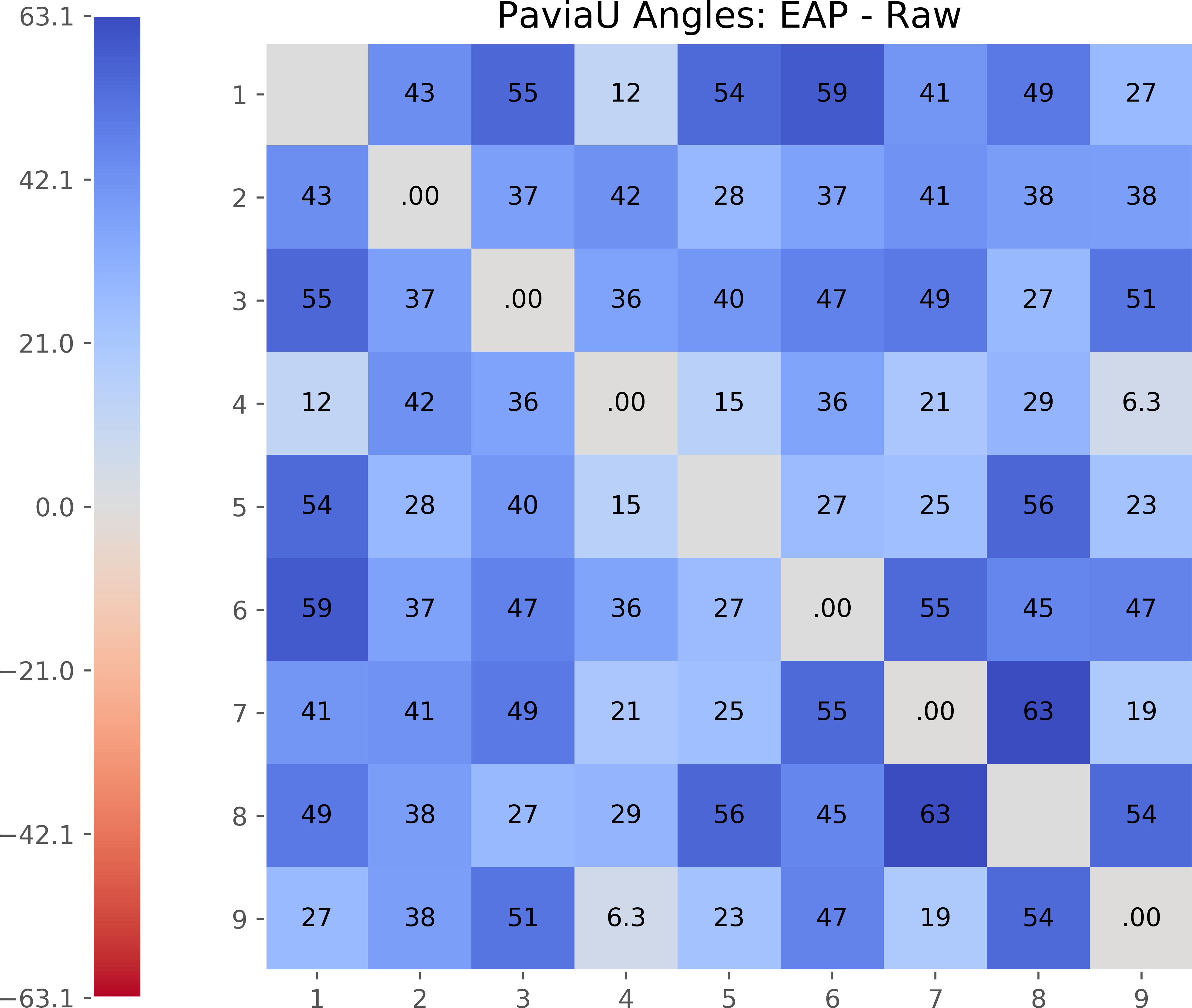}
    \includegraphics[width=0.24\textwidth]{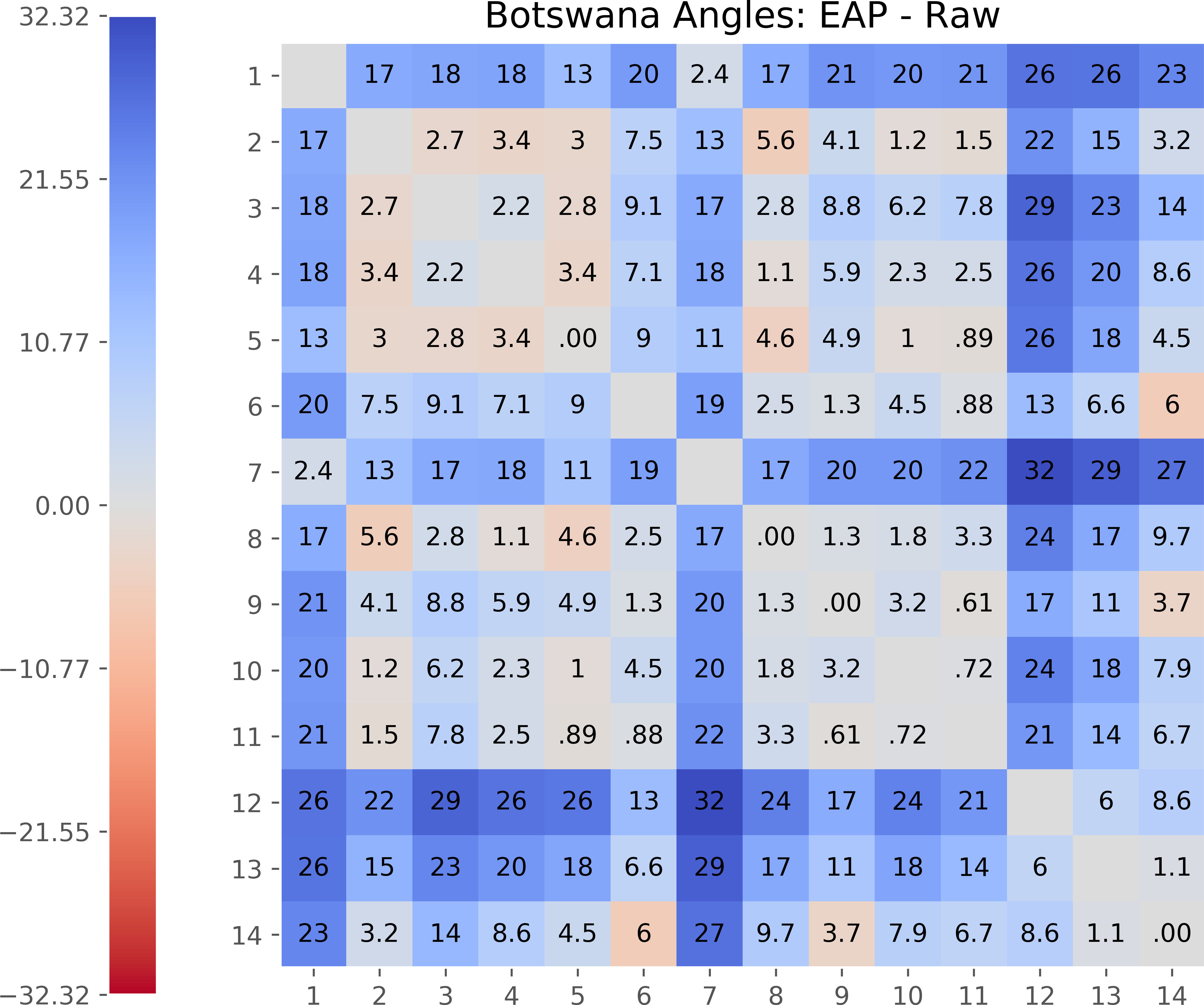} 
    \caption{Top row: class mean angles for raw features. Middle row: 3DFST minus raw. Bottom row: EAP minus raw.}
    \label{fig:angles}
\end{figure*}
\vskip-0.2cm
\section{Geometry of class means}
\label{sec:classmean}
\vskip-0.2cm
HSI features live in a high-dimensional domain and their geometric arrangement carries important information that affects classification performance. In our quest to explore such a wild landscape, we first investigate the behavior of features near their class means, which can be viewed as the simplest thing to examine, but not necessarily the most illuminating or comprehensive one. More formally, the mean $\m_j$ of the $j$-th class is simply the average of all features $\x_k$ belonging to that class. To measure angles between class means, we first subtract the center of all class means (found by averaging all $\m_j$) from each class mean $\m_j$, and use the usual definition for the angle between two vectors, $\cos\theta(\u,\v):=(\u\cdot\v)/(\|\u\| \|\v\|)$. 

Raw class mean distances and their changes are shown in Figure \ref{fig:dist}. Compared to distances between class means of raw data, 3DFST expands the distances for most pairs of classes on most datasets. There are several instances, such as in KSC, where the transform reduces the distances between class means. On the other hand, EAP being a trained neural network, is more expressive and significantly alters the arrangement of the class means. The results on KSC are visibly different from the others, primarily because classes 1--7 have means that are all tightly clustered together. 

Raw class mean angles and their changes are shown in Figure \ref{fig:angles}. We see that for most pairs of classes, 3DFST expands class mean angles. Consistent with our earlier findings on class mean changes, EAP results in larger changes to the angles between class means. The results on KSC can be explained by the raw features of classes 1--7 being tightly clustered together. Interestingly though, rather than expand the angles between these seven classes, EAP contracts the angles between the remaining classes. 

\begin{table*}[t]
	\centering
	\begin{tabular}{|c|c|c|c|c|} \hline  
		&Indian Pines &KSC &PaviaU &Botswana \\ \hline 
		Raw  &$0.0551  \pm  0.0347$ &$0.0722 \pm 0.0616$ &$0.0465 \pm  0.0377$ &$0.0283  \pm 0.0202$ \\ \hline 
		3DFST &$0.0817 \pm  0.0372$ &$0.0629   \pm 0.0469$ &$0.0714 \pm 0.0522$ &$0.0406  \pm 0.0281$ \\ \hline 
		EAP  &$0.1735 \pm 0.0449$ &$0.1774  \pm  0.0246$ &$0.1394 \pm 0.0752$ &$0.1659  \pm  0.0371$ \\ \hline
	\end{tabular}
	\caption{Average class variability $\pm$ standard deviations}
	\label{tab:radius}
\end{table*}
\vskip-0.2cm
\begin{figure*}[t]
	\centering
	\begin{subfigure}[b]{0.24\textwidth}
		\includegraphics[width=\textwidth]{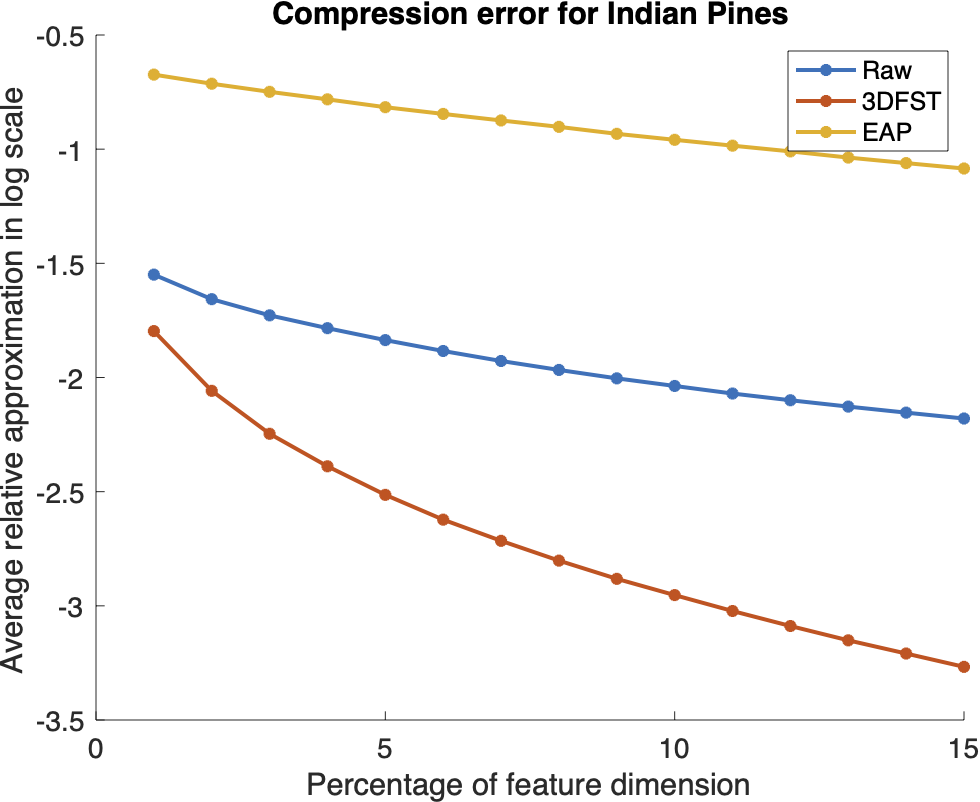}
		\caption{Indian Pines}
	\end{subfigure}
	\begin{subfigure}[b]{0.24\textwidth}
		\includegraphics[width=\textwidth]{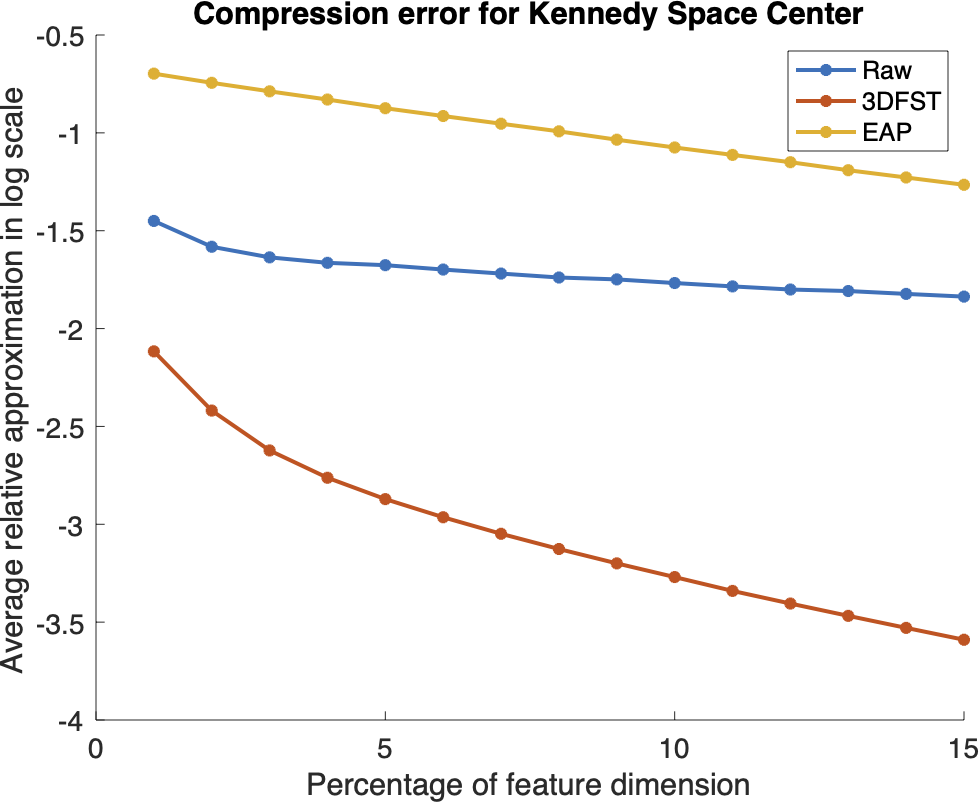}
		\caption{Kennedy Space Center}
	\end{subfigure}
	\begin{subfigure}[b]{0.24\textwidth}
		\includegraphics[width=\textwidth]{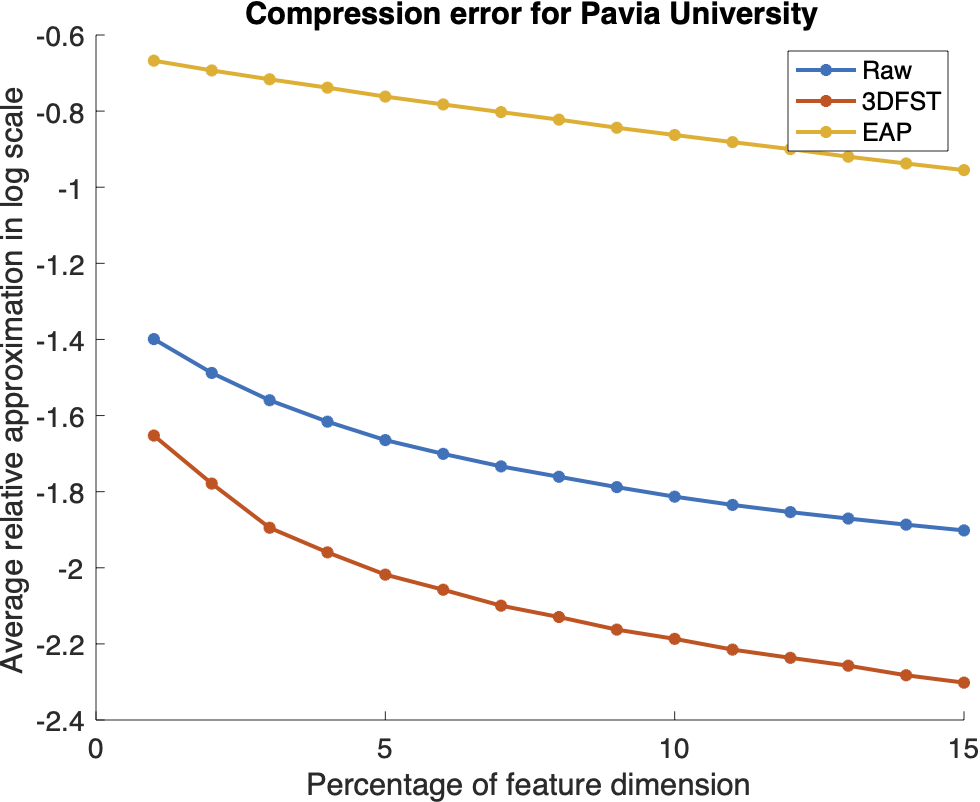}
		\caption{Pavia University}
	\end{subfigure}
	\begin{subfigure}[b]{0.24\textwidth}
		\includegraphics[width=\textwidth]{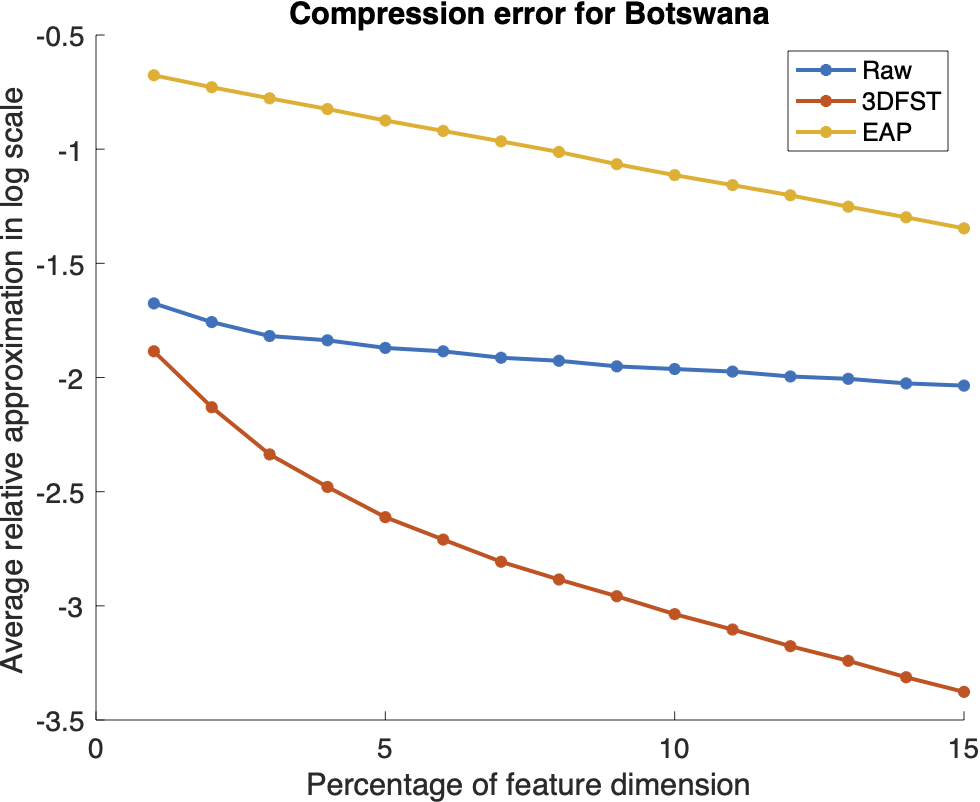}
		\caption{Botswana}
	\end{subfigure}
	\caption{Average approximation error}
	\label{fig:compress}
\end{figure*}
\begin{figure*}[!]
	\centering
	\includegraphics[width=0.24\textwidth]{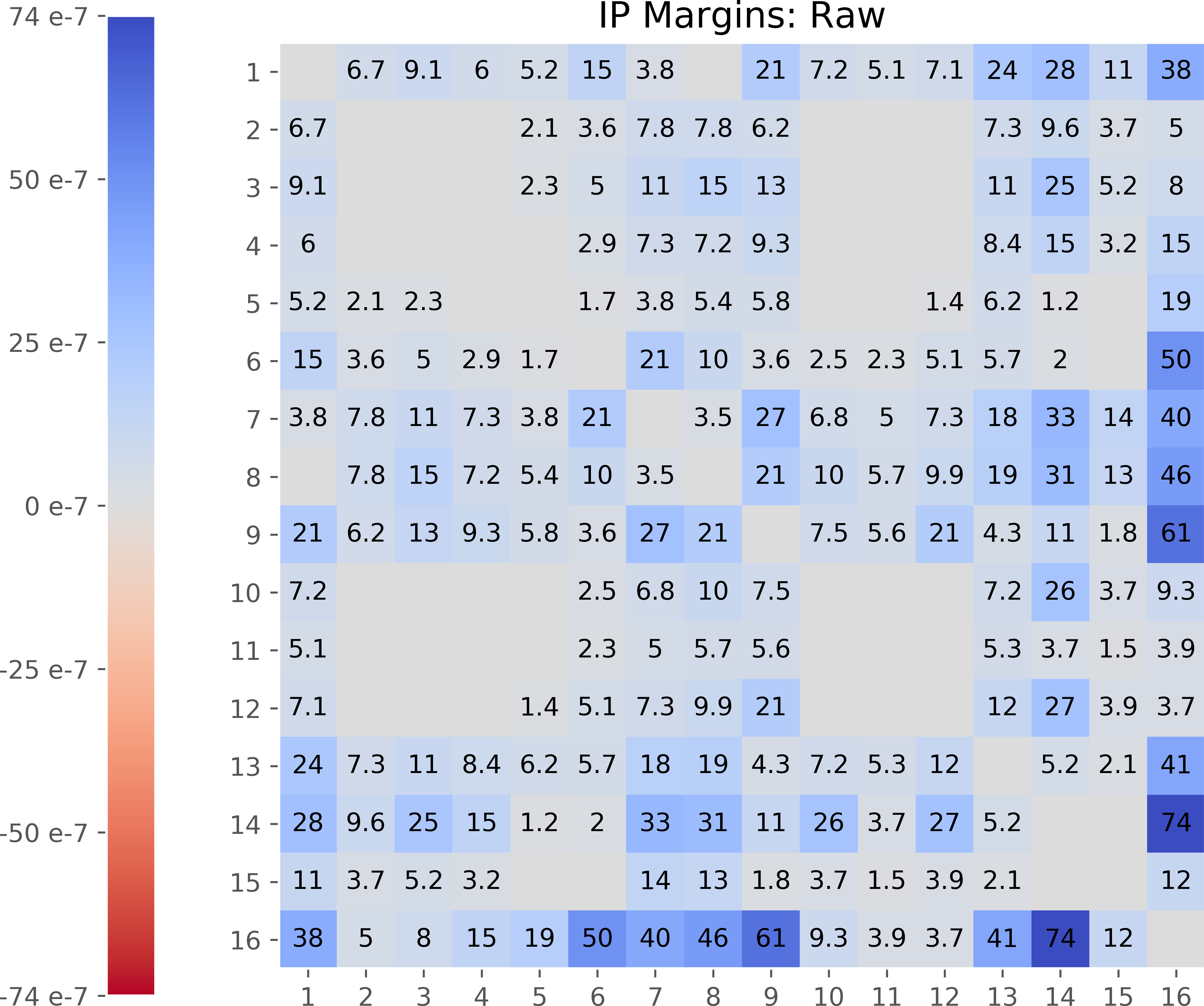}
	\includegraphics[width=0.24\textwidth]{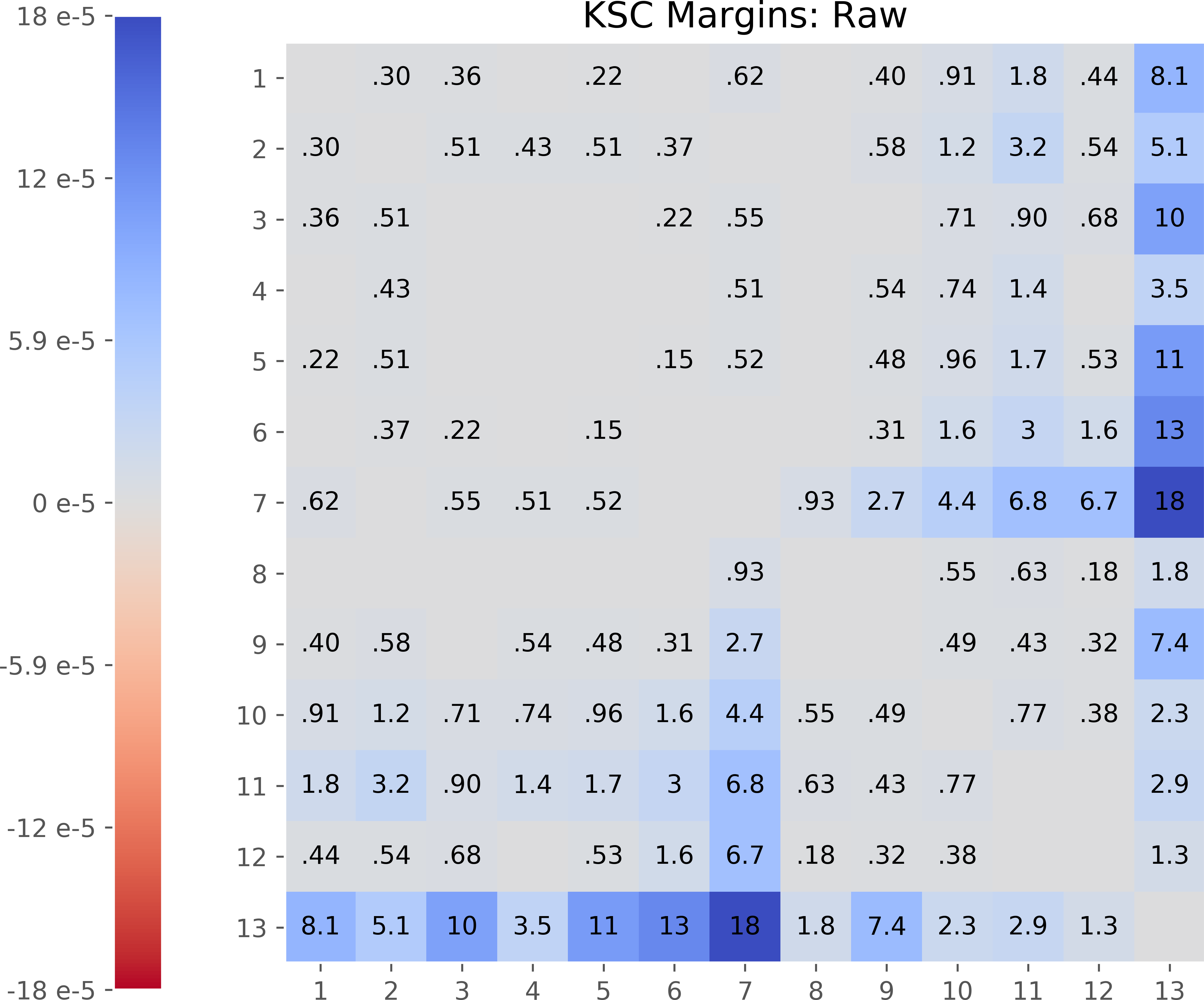}
	\includegraphics[width=0.24\textwidth]{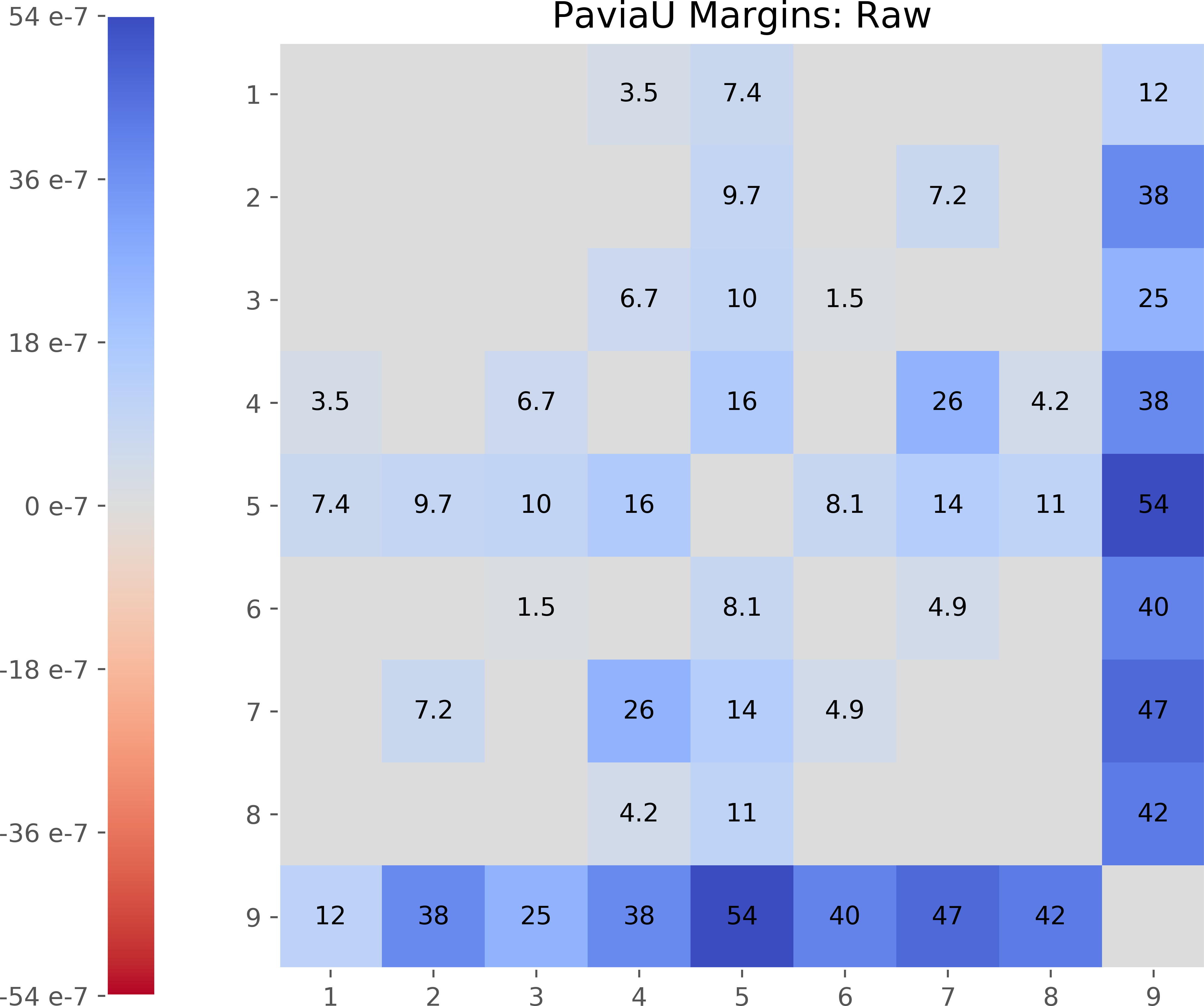}
	\includegraphics[width=0.24\textwidth]{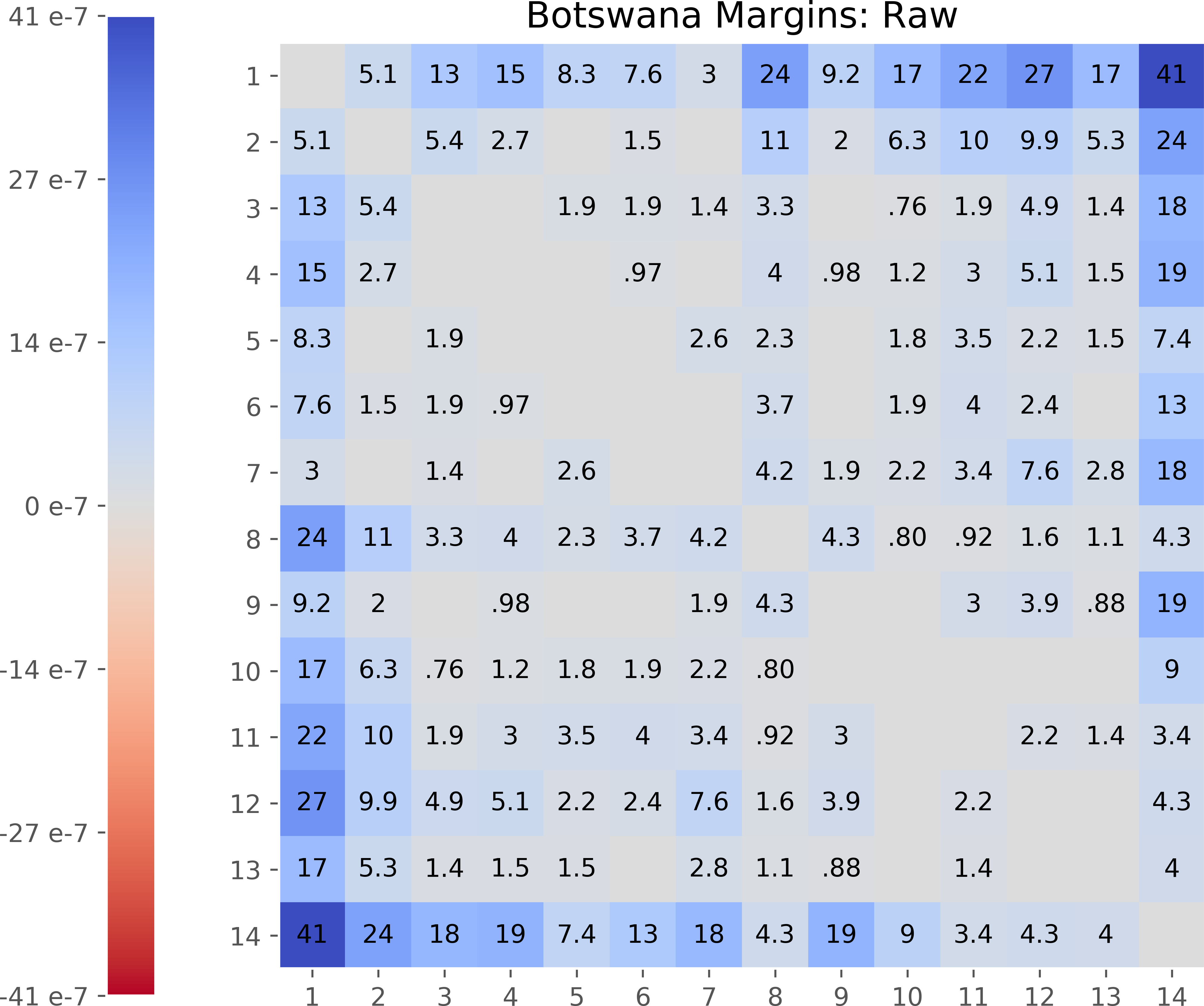} \\
	\includegraphics[width=0.24\textwidth]{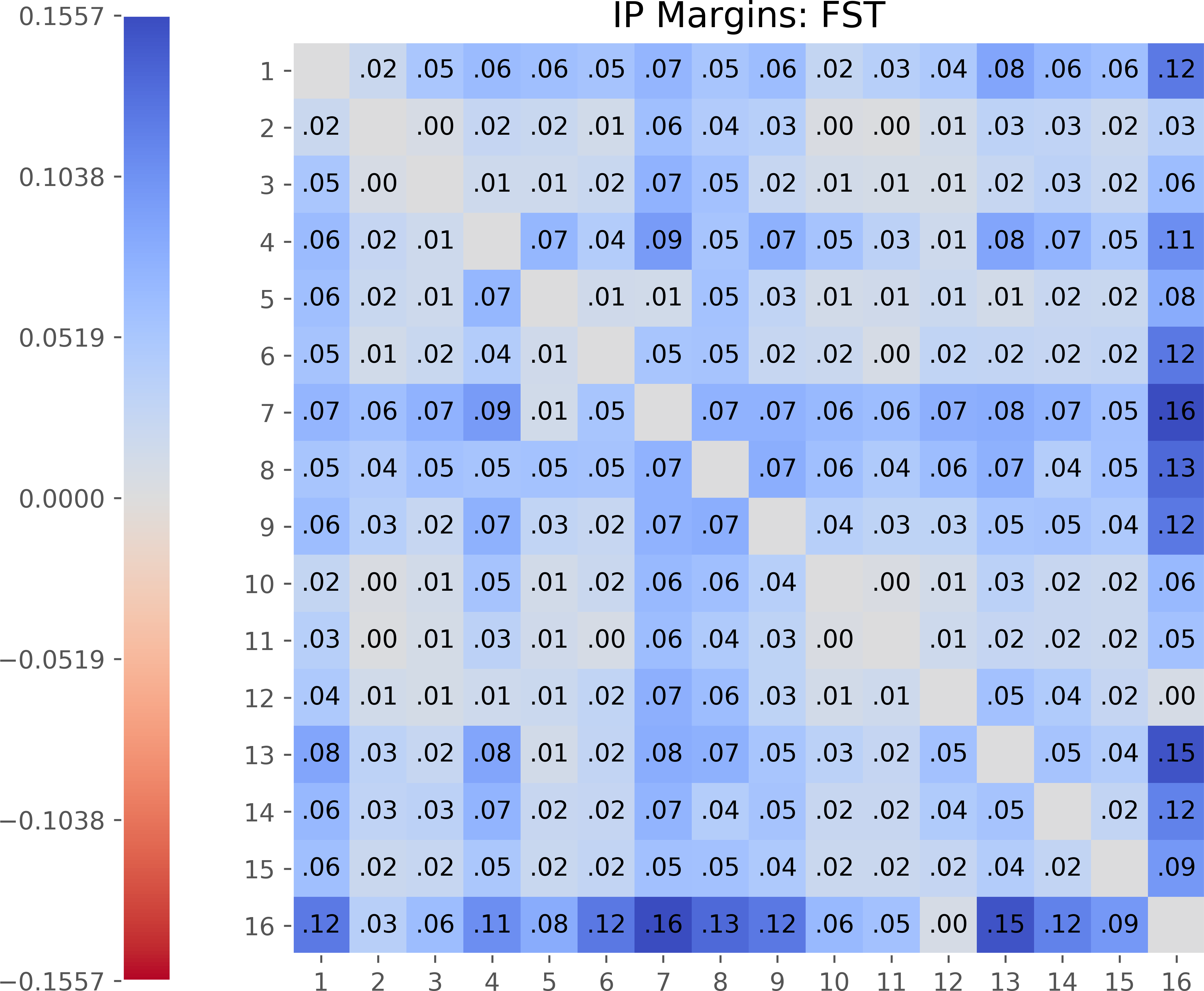}
	\includegraphics[width=0.24\textwidth]{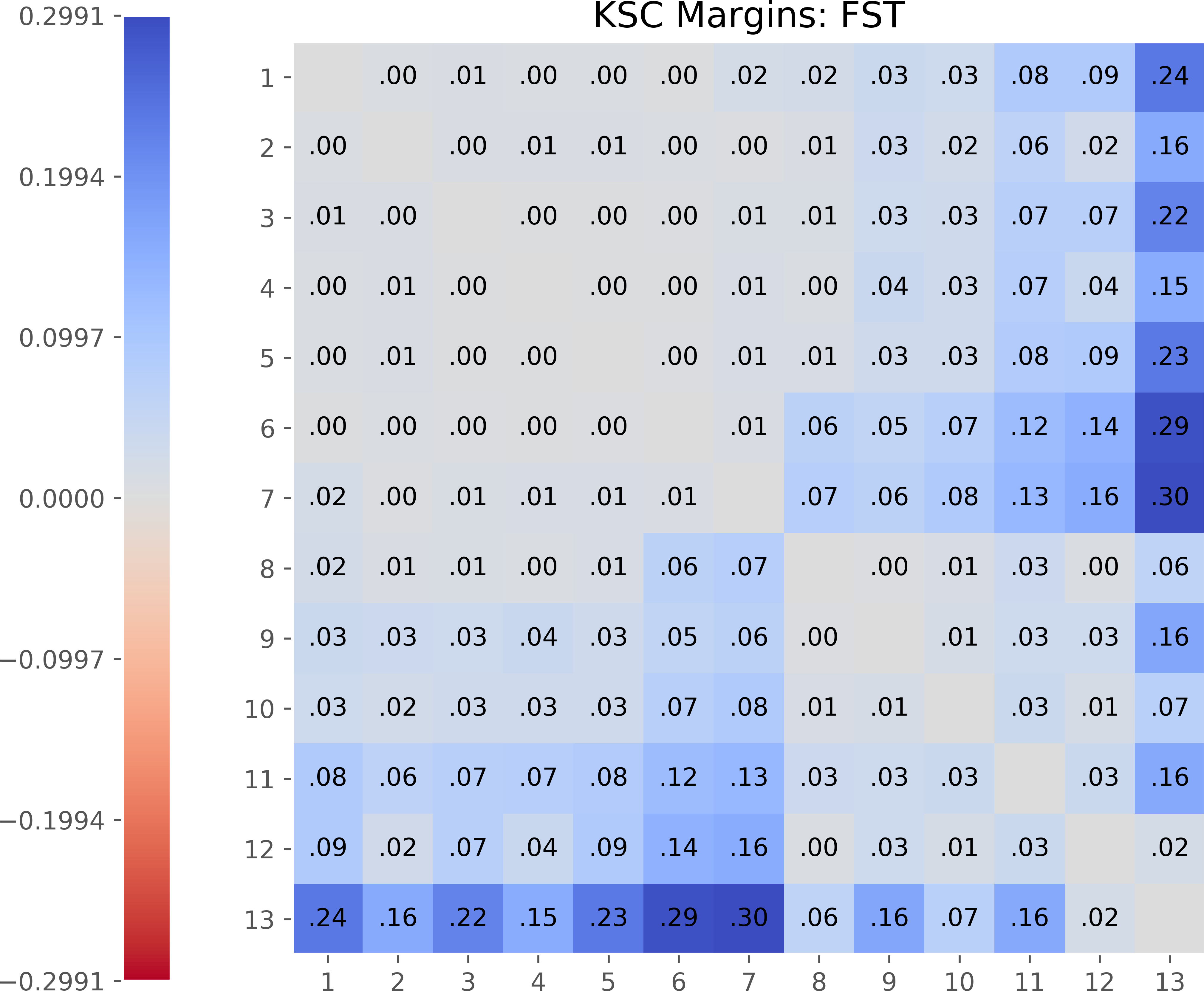}
	\includegraphics[width=0.24\textwidth]{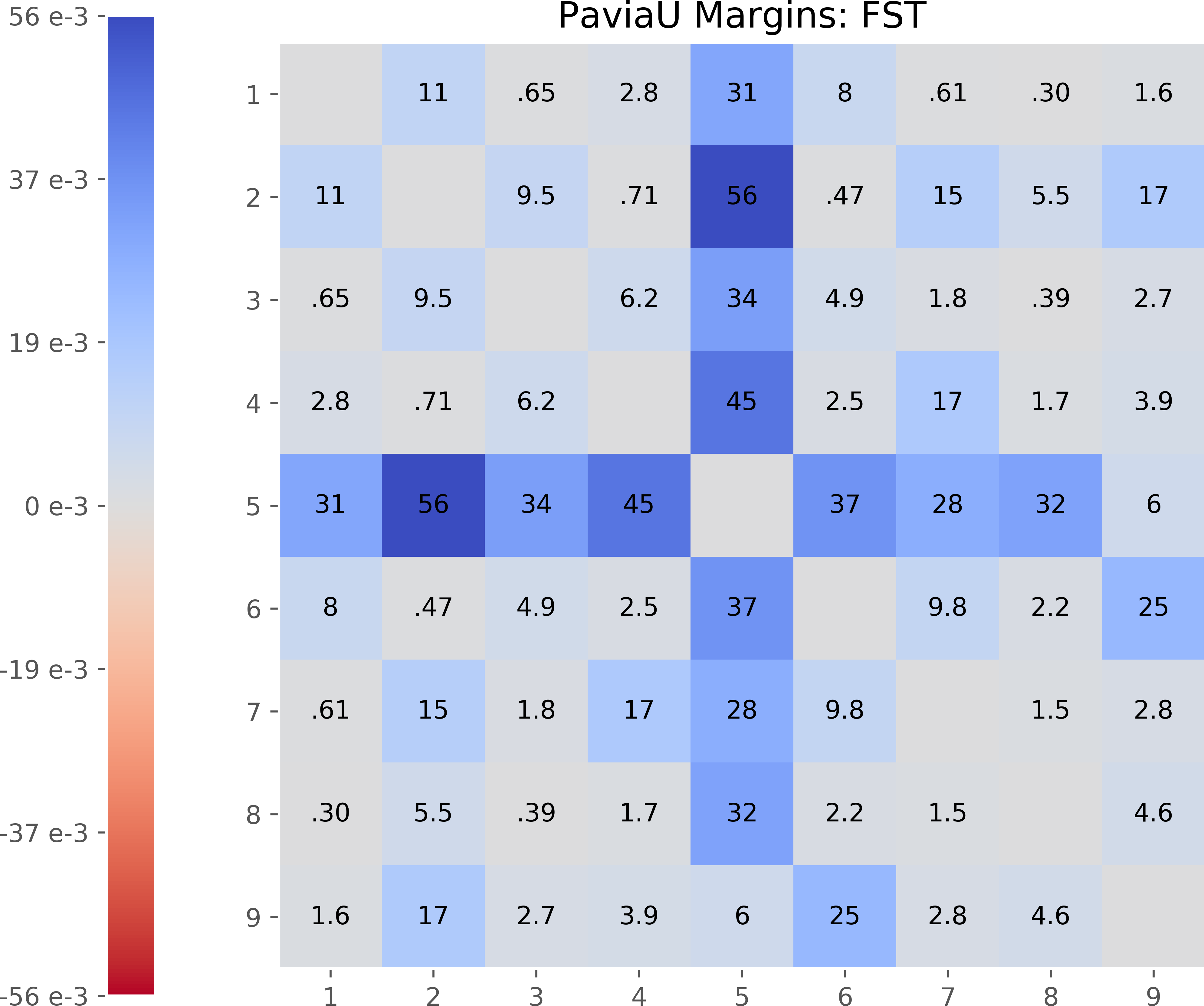}
	\includegraphics[width=0.24\textwidth]{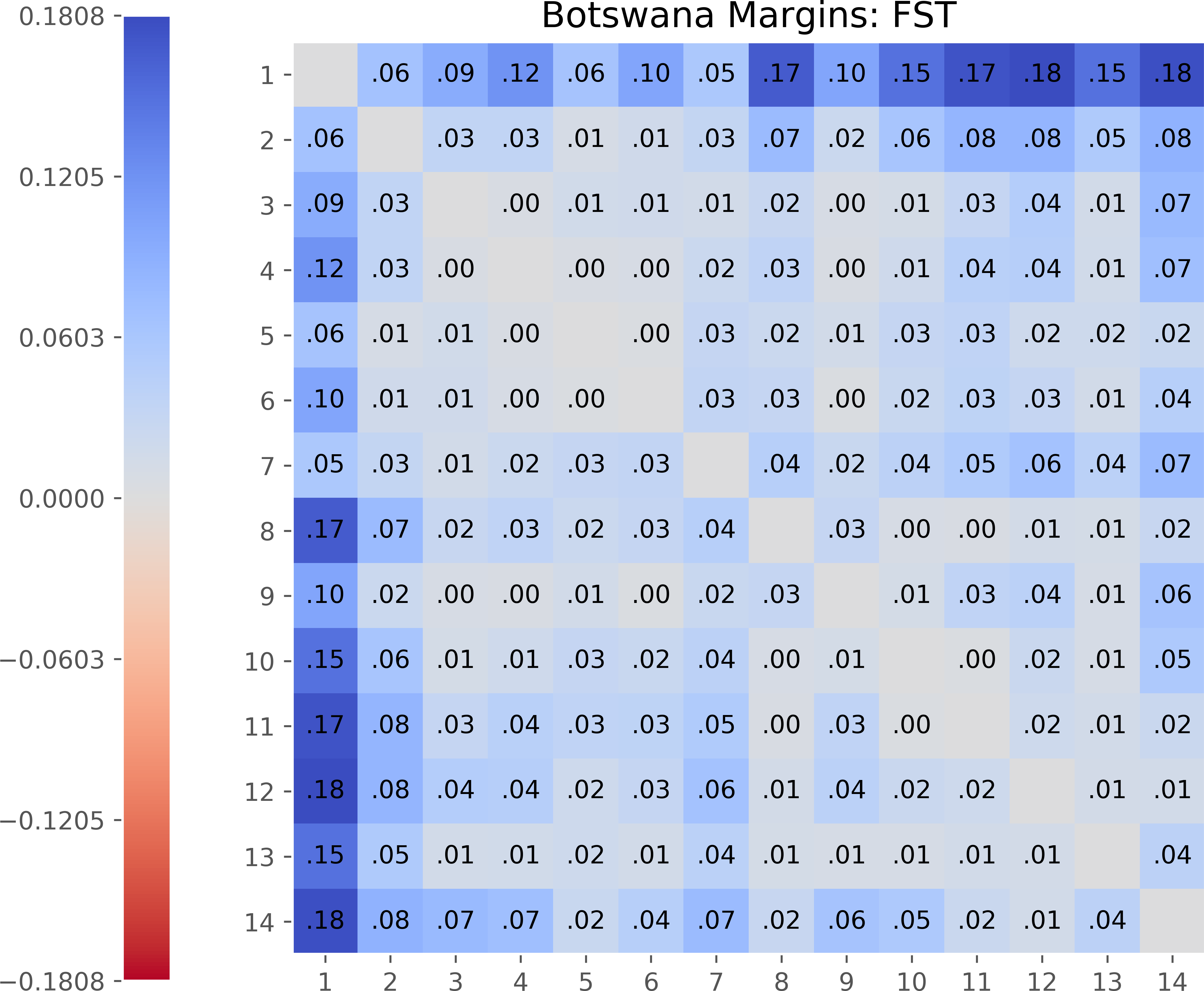}
	\caption{Maximum margins between classes. Top row: raw features. Bottom row: 3DFST features. Blank spots in the top row indicate class pairs that are not linearly separable.}
	\label{fig:sep}
\end{figure*}
\vskip-0.2cm
\begin{table*}[!]
	\centering
	\begin{tabular}{|c|c|c|c|c|} \hline  
		&Indian Pines &KSC &PaviaU &Botswana  \\ \hline 
		Neural collapse angle &$93.8226$  &$94.7802$ &$97.1808$ &$94.4117$ \\ \hline
		EAP class distances &$0.2488 \pm   0.1281$ &$0.1015 \pm 0.0614$ &$0.3710 \pm   0.1358$ &$0.2294  \pm  0.1342$ \\ \hline   
		EAP class angles & $25.9985 \pm 14.9650$ &$6.0478  \pm  3.3516$ &$61.7212\pm 12.3580$ &$21.2320 \pm 15.7333$ \\ \hline
	\end{tabular}
	\caption{Average distances and angles between classes (in degrees) $\pm$ standard deviations}
	\label{tab:neuralcollapse}
\end{table*}

\vskip-0.2cm
\section{Compression and low-dimensionality}
\vskip-0.2cm
The geometry of class means provides a first-order statistical description of the high-dimensional geometry of HSI features. Intra-class variability can be quantified by second order statistics such as squared deviation away from the mean. Table \ref{tab:radius} displays the average Euclidean distance between feature vectors and their means. Across all four data sets, both feature transforms result in significant changes to the average radius of each class. However, EAP expands the average class radius by a significantly greater amount than 3DFST. This illustrates an important trade-off: while classes are generally mapped further apart and with greater angles by EAP, the cluster sizes themselves appear to expand. 

Our next set of experiments explores whether class features are isotropic (look like balls) or anisotropic (look like ellipsoids). To do this, we approximate each class with the best possible low-dimensional hyperplane, whose dimension is variable. The total approximation or compression error measured in the Euclidean norm decreases as the hyperplane dimension increases. This provides insight into class geometries: for anisotropic classes, the error decreases faster than that of isotropic ones. Our results summarized in Figure \ref{fig:compress} indicate an important distinction between the tendencies of 3DFST and EAP: the former creates anisotropic features relative to the feature space dimension, whereas the latter creates more isotropic classes. In fact, our results indicate that EAP features are more isotropic than the raw spectra.

\vskip-0.2cm
\section{3DFST and linear separability}
\vskip-0.2cm
Our empirical observations indicate that 3DFST takes a more conservative approach. Rather than transform classes across large distances, it contracts variability along the normal directions This has significant on the linear separability between classes. As shown in Figure \ref{fig:sep}, even if raw spectra belonging to different classes are linearly separable, their distances to the maximum margin hyper-plane is small and on the order of $10^{-4}$. On the other hand, for every pair of classes in each dataset, their 3DFST features are linearly separable by hyperplanes whose margins are orders of magnitudes larger. The margin of a hyper-plane has important implications to the generalization error when a subset of the data points are used to determine a hyperplane split \cite{vapnik1998statistical,bartlett1999generalization}. 

\vskip-0.2cm
\section{EAP and Neural Collapse?}
\vskip-0.2cm
It is natural to wonder if EAP induces neural collapse or similar. While there are many parallels, such as EAP generally amplifying the distances and angles between class means, we point out several inconsistencies. In neural collapse, the distances between class means would be constant and the class means (after re-centering) would form a (scaled and rotated) simplex ETF. The angles between vectors in neural collapse are $\cos^{-1}(-\frac{1}{m-1})$, where $m$ is the number of classes. For the four datasets used in this article, Table \ref{tab:neuralcollapse} summarizes the distances and angles between classes for EAP features (recall that the class means are shifted by the center of these means). These numbers are not consistent with what qualifies as neural collapse.

\vskip-0.2cm
\section{Insights and Discussion}
\label{sec:typestyle}
\vskip-0.2cm
In this paper we examined features generated by two state-of-the-art algorithms: 3DFST and EAP. We have demonstrated that they operate on drastically different geometric principles even though they yield comparable classification results. The results in this article naturally raises two questions. 

The first question is what is the desired architecture for processing HSI? Compared with the state-of-the-art image classification networks, the ones considered here are rather shallow. The second question is whether neural collapse holds for other distributions, like those present in the HSI setting? We saw that both features extractors, while significantly different from each other, do not generate features consistent with neural collapse. 
\vskip-0.2cm
\bibliographystyle{IEEEbib}
\bibliography{refs}

\end{document}